\documentclass{article}

    \PassOptionsToPackage{numbers, compress}{natbib}

\usepackage[preprint]{neurips_2025}

\usepackage[utf8]{inputenc} 
\usepackage[T1]{fontenc}    
\usepackage{hyperref}       
\usepackage{url}            
\usepackage{booktabs}       
\usepackage{amsfonts}       
\usepackage{nicefrac}       
\usepackage{microtype}      
\usepackage{xcolor}         
\usepackage{amsmath}
\usepackage{enumitem}

\usepackage{graphicx}
\usepackage{multirow}
\usepackage{float}

\title{Universal Few-Shot Spatial Control \\for Diffusion Models}

\author{
    \textbf{Kiet T.~Nguyen~~~~\ Chanhyuk Lee~~~~\ Donggyun Kim~~~~\ Dong Hoon Lee~~~~\ Seunghoon Hong} \\
    KAIST \\
    \footnotesize{
    \texttt{\{kietngt00, chan3684, kdgyun425, donghoonlee, seunghoon.hong\}@kaist.ac.kr}
    }
}

\begin{document}

\maketitle
\vspace{-0.1in}

\begin{figure}[!h]
  \centering
  \includegraphics[width=0.98\linewidth]{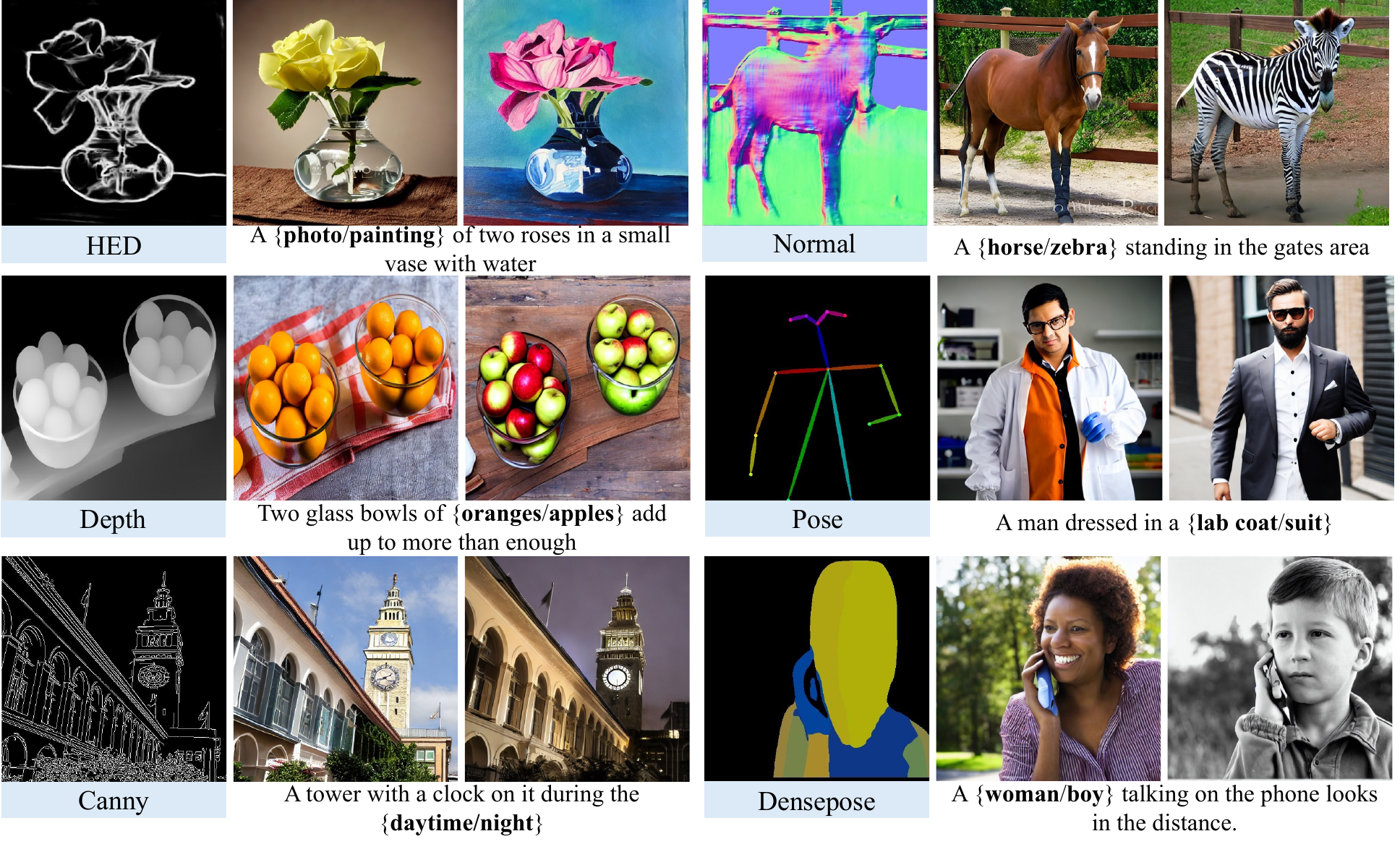}
  \vspace{-0.15in}
  \caption{Results of our method learned with \textbf{30 examples} on \textbf{unseen} spatial conditions. The proposed control adapter guides the pre-trained T2I models in a versatile and data-efficient manner. }
    \vspace{-0.05in}
\end{figure}

\begin{abstract}
  Spatial conditioning in pretrained text-to-image diffusion models has significantly improved fine-grained control over the structure of generated images.
  However, existing control adapters exhibit limited adaptability and incur high training costs when encountering novel spatial control conditions that differ substantially from the training tasks.
  To address this limitation, we propose Universal Few-Shot Control (UFC), a versatile few-shot control adapter capable of generalizing to novel spatial conditions.
  Given a few image-condition pairs of an unseen task and a query condition, UFC leverages the analogy between query and support conditions to construct task-specific control features, instantiated by a matching mechanism and an update on a small set of task-specific parameters. 
  Experiments on six novel spatial control tasks show that UFC, fine-tuned with only 30 annotated examples of novel tasks, achieves fine-grained control consistent with the spatial conditions.
  Notably, when fine-tuned with 0.1\% of the full training data, UFC achieves competitive performance with the fully supervised baselines in various control tasks. 
  We also show that UFC is applicable agnostically to various diffusion backbones and demonstrate its effectiveness on both UNet and DiT architectures. Code is available at \url{https://github.com/kietngt00/UFC}.

\end{abstract}


\section{Introduction}

Text-to-Image (T2I) diffusion models~\cite{sdv3, sdv15, podell2024sdxl, saharia2022photorealistic, chen2024pixart, chen2024pixartdelta, flux2024}, trained on large-scale datasets, have achieved remarkable success in generating high-quality, semantically aligned images from natural language prompts. While language-based control offers intuitive and flexible guidance, it often lacks the precision needed for fine-grained visual control, such as specific object positions, shapes, or scene layouts. To overcome this, recent works~\cite{lhhuang2023composer,t2iadapter,li2023gligen,zhang2023adding,controlnetpp,unicontrol,zhao2023uni,wang2023incontext} incorporate explicit spatial signals—like edge maps, depth maps, and segmentation masks to control diffusion models.

To enable spatial control while preserving the generative quality of pre-trained diffusion models, existing methods typically employ control adapters~\cite{zhang2023adding,t2iadapter,li2023gligen} that inject spatial signals into a frozen T2I model. However, these adapters are usually trained independently for each spatial control task, requiring substantial computational resources and extensive labeled data for a new task. Alternatively, reusing pre-trained multi-task adapters - either directly~\cite{unicontrol,wang2023incontext} or with minimal updates~\cite{zhao2023uni}- struggle to generalize to tasks that differ from their training distribution, and often show poor adaptability.

In this work, we aim to address these limitations by proposing a universal few-shot learning framework, named Universal Few-shot Control (UFC), that efficiently controls diffusion models with novel spatial conditions using only a small number of labeled examples at test time (\emph{e.g.}, dozens of image and condition pairs). 
Developing such a method would substantially enhance the practicality and flexibility of spatial control methods, but also poses two main challenges. First, significant domain discrepancies across diverse spatial signals make it difficult to learn a consistent, shared representation of control signals that facilitate generalization to novel conditions. 
Second, in real-world settings where new control tasks may arise dynamically and labeled data is limited, the model must be capable of efficient test-time adaptation from only a few annotated examples.

To overcome these challenges, UFC introduces a universal control adapter that represents novel spatial conditions by adapting the interpolation of visual features of images in a small support set, rather than directly encoding task-specific conditions. 
The interpolation is guided by patch-wise similarity scores between the query and support conditions, modeled by a matching module~\cite{kim2023universal}. 
Since image features are inherently task-agnostic, this interpolation-based approach naturally provides a unified representation, enabling effective adaptation across diverse spatial tasks. Furthermore, to facilitate rapid and data-efficient updates, UFC combines episodic meta-learning on a multi-task dataset with a small set of task-specific parameters, enabling effective adaptation under few-shot settings. 

Extensive experiments across diverse spatial control modalities, including edge maps, depth maps, normal maps, human pose, and segmentation masks for human body, demonstrate that UFC outperforms prior approaches in few-shot settings by large margins. 
Notably, UFC achieves fine-grained spatial control using only \textbf{30 shots}, and even on par with some fully supervised baselines on Normal, Depth, and Canny with just 150 support examples (0.1\% of the training data). Furthermore, UFC is compatible with both UNet~\cite{unet} and DiT~\cite{dit} backbones, ensuring applicability across recent T2I diffusion architectures.
To our knowledge, we propose the first method for \emph{few-shot spatial control} in text-to-image diffusion models, which enables spatial control image generation with novel conditions at test time using minimal annotated data.
\vspace{-0.05in}
\section{Related Works}
\vspace{-0.05in}
\paragraph{Controllable Image Diffusion Models}
Text-to-Image diffusion models~\cite{sdv3, sdv15, podell2024sdxl, saharia2022photorealistic, chen2024pixart, chen2024pixartdelta, flux2024}
have gained widespread adoption for their ability to generate high-quality, semantically aligned images from text prompts. 
To enable finer control over image structure, various methods~\cite{t2iadapter, zhang2023adding, li2023gligen, controlnetpp, unicontrol, zhao2023uni, wang2023incontext, tan2024ominicontrol, wang2025unicombine} proposed incorporating additional spatial control inputs such as edge maps, segmentation masks, or human poses, etc., into diffusion models. ControlNet~\cite{zhang2023adding} augments a pretrained T2I diffusion model with a control adapter, initialized from the model itself, to encode spatial conditions and inject them into the frozen diffusion model. Uni-ControlNet~\cite{zhao2023uni} extends this by adding lightweight feature extractors to support multiple condition types. 
Prompt Diffusion~\cite{wang2023incontext}, built on ControlNet, introduces a vision-language prompting mechanism to enable in-context learning for handling a diverse set of tasks.
However, these prior methods are limited to tasks seen during training. 
To address this limitation, training-free methods~\cite{ugd, yu2023freedom, mo2024freecontrol, ctrl_x} have been proposed; but, they often rely on a long process of latent optimization, which substantially increases generation time while providing only limited controllability. Consequently, few-shot adaptation with more fine-grain controllability without incurring significant generation-time overhead remains unexplored.

\vspace{-0.1in}
\paragraph{Few-shot Diffusion Models} 
To extend diffusion models to unseen tasks, several works~\cite{cao2024few, zhu2022few, yang2024few, sinha2021few, giannone2022few, Li_2023_ICCV_few, jin2025dualinterrelateddiffusionmodelfewshot} have explored few-shot learning for image generation. For example, D2C~\cite{sinha2021few} trains a variational autoencoder~\cite{vae} on a few labeled examples to produce latent representations that condition a diffusion model to generate images in the target domain. Similarly, FSDM~\cite{giannone2022few} uses visual features extracted from a small support set to guide image generation toward unseen classes. While these approaches demonstrate promising few-shot generation capabilities, they primarily focus on domain or class transfer, and are not designed to generalize across diverse spatial control tasks.

\vspace{-0.1in}
\paragraph{Analogy Image Generation.} 
Several recent works~\cite{gu2024analogist, yang2023imagebrush} leverage the analogy between a query source and a support source-target pair to enable generalization in diffusion models on diverse tasks with visual instruction.
Analogist~\cite{gu2024analogist} treats visual instruction as an inpainting task, arranging query-support pairs in a $2 \times 2$ grid and using a diffusion model to complete the missing cell. This training-free method generalizes well to tasks like colorization, deblurring, and editing. While we also leverage query-support analogy in few-shot settings, our goal differs: prior work preserves the support image’s appearance, whereas we aim to generate diverse content guided by spatial conditions.
\section{Background}
\label{sec:preliminaries}
\vspace{-0.05in}

A text-to-image diffusion model generates an image $x$ conditioned on a text prompt $c_\text{text}$ by iteratively denoising a sequence of latent variables.
Specifically, the model learns a finite-length Markov chain $\{z_{T-t}\}_{t=0}^T$ that starts from standard Gaussian noise $z_T \sim \mathcal{N}(0, I)$, where each transition $z_{t+1} \to z_t$ progressively removes noise.
The final clean variable $z_0$ can be either the image itself or a latent representation $z = E(x)$, which is decoded back to the image by $x = D(z)$, where $(E, D)$ denotes an auto-encoder~\cite{sdv15,podell2024sdxl}.
The chain is the time-reversal of a diffusion process $\{z_t\}_{t=0}^T$ that gradually corrupts the data by adding small Gaussian noise until it becomes pure noise at $z_T$~\cite{sohl2015deep}.
To learn each denoising transition, the diffusion network $\mathcal{E}_\phi$ is trained to predict the added noise $\epsilon$ at timestep $t$:
\vspace{-0.05in}
\begin{equation}
    \min_\phi~ \mathbb{E}_{(z_0, c_\text{text})\sim \mathcal{D}_\text{train},t,\epsilon} \left[ \left\| \epsilon -  \mathcal{E}_\phi(z_t, c_\text{text}, t) \right\|^2 \right],
\end{equation}

\vspace{-0.1in}
where ($z_0, c_\text{text}$) is drawn from the training dataset $\mathcal{D}_\text{train}$ and $t$ is uniformly sampled from $\{1, \cdots, T\}$.
$z_t$ can be computed in closed form from $z_0$ and $t$ using the properties of Gaussian variables~\cite{ho2020denoising}.
After training, an image is generated by running the learned Markov chain from $z_T$ down to $z_0$.

Although text-to-image diffusion models control the overall content of an image, a text prompt alone is limited in steering the fine-grained spatial structures.
Spatial control approaches~\cite{t2iadapter, zhang2023adding, li2023gligen, tan2024ominicontrol} address this limitation by introducing an additional spatial condition $y_\tau$, typically a dense map describing the target image (\emph{e.g.,} edges, depth, or semantic masks).
For each condition type (or \emph{task}) $\tau$, an auxiliary control adapter $\mathcal{G}_{\theta_\tau}(y_\tau)$\footnote{The control adapter can take auxiliary inputs such as noisy latent, timestep, and text prompt, but we omit them for clarity.} is trained and injected into a frozen, pre-trained diffusion backbone $\mathcal{E}_\phi$, while keeping the backbone parameters $\phi$ fixed to preserve their learned priors: 
\vspace{-0.07in}
\begin{equation}
    \min_{\theta_\tau}~ \mathbb{E}_{(z_0, c_\text{text}, y_\tau) \sim \mathcal{D}_\text{train},t,\epsilon} \left[ \left\| \epsilon -  \mathcal{E}_\phi(z_t, c_\text{text}, t, \mathcal{G}_{\theta_\tau}(y_\tau)) \right\|^2 \right],
    \label{eq:spatial_control_denoising}
\end{equation}

\vspace{-0.09in}
However, learning a dedicated adapter for each task $\tau$ usually requires thousands of labeled data to achieve reasonable performance~\cite{zhang2023adding}.
This can be costly if we want to condition the pre-trained model on a novel task $\tau_\text{novel}$.
On the other hand, reusing an adapter trained on a fixed task(s) for novel tasks often suffers from limited adaptation performance, especially when the new tasks are substantially different from training~\cite{zhao2023uni}.
Therefore, designing a \emph{versatile} control adapter that can accommodate arbitrary unseen spatial conditions with a small amount of data remains a key challenge for data-efficient steering of text-to-image diffusion models.

\vspace{-0.07in}
\section{Method}
\label{sec:method}
\vspace{-0.05in}

\subsection{Problem Setup}
\label{sec:problem_setup}
\vspace{-0.05in}

Our goal is to build a universal few‑shot framework that steers a frozen text‑to‑image diffusion model with arbitrary \emph{novel} spatial conditions, given only a handful of annotated examples (typically a few dozen).
To this end, we introduce a universal control adapter $\mathcal{I}(y_\tau; \mathcal{S}_\tau)$ that can inject any type of spatial conditions into the frozen diffusion model by directly incorporating the given data as follows:
\begin{equation}
\hat{\epsilon} = \mathcal{E}_\phi(z_t, t, c_{\text{text}}, \mathcal{I}(y_\tau; \mathcal{S}_\tau)),
\label{eq:fewshot}
\end{equation}
where $\mathcal{S}_\tau = \{(x^i, y_\tau^i)\}_{i \leq N}$ denotes a support set comprising $N$ image-condition pairs for task $\tau$.
We focus exclusively on the design of the adapter $\mathcal{I}$ while the remaining components, such as loss function, diffusion architecture, and injection strategy, are indepedent to the adapter and can be freely chosen from existing spatial control approaches~\cite{t2iadapter,zhang2023adding,li2023gligen}.

Unlike a task-specific adapter $\mathcal{G}_{\theta_\tau}$ (Eq.~\eqref{eq:spatial_control_denoising}), the universal adapter $\mathcal{I}$ must address two key challenges.
\begin{enumerate}[leftmargin=0.5cm]
    \item \textbf{Heterogeneous input formats.}
    Spatial conditions vary widely—from extremely sparse cues such as human‑pose keypoints to dense maps such as depth or segmentation.
    The adapter must transform these disparate inputs into the \emph{unified} control features that the frozen diffusion model can consume reliably.
    
    \item \textbf{Severe data scarcity.}
    For a new task, we assume only a few dozen labeled examples. To handle distribution shifts without over‑fitting the tiny support set, the adapter needs an adaptation mechanism that is both \emph{flexible} and \emph{efficient}.
\end{enumerate}

\subsection{Universal Few-Shot Control Adapter}
\label{sec:adapter}

\begin{figure}[!t]
  \centering
  \includegraphics[width=\linewidth]{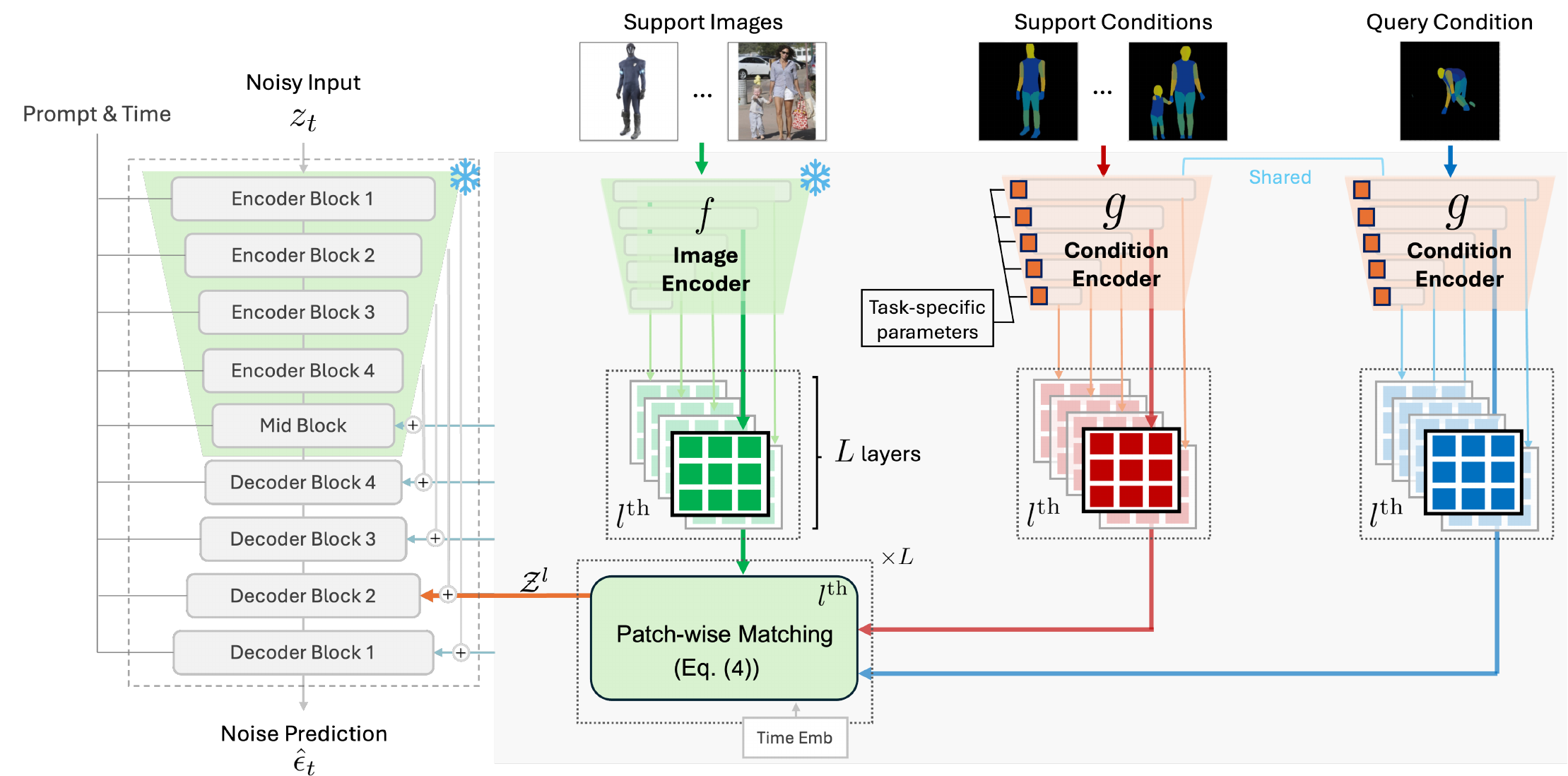}
  \vspace{-0.25in}
  \caption{
    Overview of the proposed framework. The control adapter $\mathcal{I}$ consists of an image encoder $f$, a condition encoder $g_\tau$, and a matching module implementing Eq.~\ref{eq:matching}. The support image-condition pairs and the query conditions are encoded to extract multi-layer features. The matching module at each layer is applied to produce control features. The features are then injected into the generation process following the mechanism in Section~\ref{sec:architecture} to control the structure of images.
    }
    \label{fig:arch}
    \vspace{-0.1in}
\end{figure}

We propose \textbf{Universal Few-shot Control (UFC)}, a universal control adapter that addresses the challenges outlined in Section \ref{sec:problem_setup} by (1) unifying heterogeneous spatial conditions with image features, and (2) adapting to new tasks through patch-wise matching and parameter-efficient fine-tuning.
An overview is shown in Figure~\ref{fig:arch}.


To obtain control features that remain consistent across diverse condition types, UFC leverages \emph{task-agnostic} visual patches extracted from support images via \textbf{patch-wise matching}.
Specifically, given a query condition $y^q_\tau$ and a support set $\mathcal{S}_\tau = \{(x^i, y^i_\tau)\}_{i \leq N}$, UFC encodes images and conditions into spatial feature maps with encoders $f$ and $g_\tau$, respectively.
The control feature for $k$-th query patch is then constructed by
\begin{equation}
\mathcal{I}(y^{q,k}_\tau; \mathcal{S}_\tau) = \sum_{i=1}^N \sum_{j=1}^M \sigma\big(g_\tau(y^{q,k}_\tau),\, g_\tau(y_\tau^{i, j})\big) \cdot f(x^{i,j}),
\label{eq:matching}
\end{equation}
where $\sigma: \mathbb{R}^d \times \mathbb{R}^d \rightarrow [0, 1]$ is a patch-wise similarity function, $M$ is the number of spatial patches per image, and $k, j \le M$ denote the patch indices.
Intuitively, the patch embeddings of the support images $f(x^{i,j})$ form  \emph{task-agnostic} bases of query conditions, while \emph{task-specific} weights—computed from the conditions—select the relevant entries.

The patch-wise matching formulation in Eq.~\ref{eq:matching}, inspired by few-shot dense prediction methods~\cite{kim2023universal, kim2024chameleon}, offers several advantages for the universal control problem.
First, since the task-specific conditions are only used to determine the weights for the task-agnostic visual features, the resulting control features $\mathcal{I}(y_\tau^q; \mathcal{S}_\tau)$ reside in a unified visual feature space regardless of the task $\tau$.
This ensures consistent guidance of the diffusion model and robust generalization to unseen tasks.
Second, by composing the control features patch-wise, it can exploit the locality inductive bias on the image-condition pairs and amplify the effective number of support patches (\emph{i.e.}, bases) for each query patch.
This provides sufficient representational power to model complex spatial controls from just a few support images.
Finally, the general matching rule can wrap any control adapter $G_{\theta_\tau}$ by treating it as a condition encoder $g_\tau$ and adding an image encoder $f$.
UFC can therefore inherit advances in adapter architectures and injection mechanisms developed for spatial control.

To adapt rapidly to a new task without over‑fitting, UFC introduces only a \emph{small} set of task-specific parameters $\theta_\tau$ while sharing the remainder.
Starting from a pre-trained visual backbone $g(\cdot; \theta)$, we obtain a condition encoder
\begin{equation}
    g_{\tau}(y_\tau) = g(y_\tau; \theta, \theta_\tau),
\end{equation}
where $\theta$ is shared across tasks and $\theta_\tau$ is lightweight (\emph{e.g.,} bias~\cite{bias-tuning} or LoRA~\cite{lora} parameters).
After meta-training on a fixed task set (Section~\ref{sec:training}), $\theta$ is frozen and only $\theta_\tau$ is fine-tuned for each novel task $\tau$.
The image encoder $f$, also initialized from a pre-trained visual backbone, remains frozen throughout to preserve task-agnostic visual features.
Leveraging strong priors of large-scale visual pre-training and modern parameter-efficient tuning techniques, UFC generalizes to unseen tasks with minimal task-specific parameters.

\vspace{-0.05in}
\subsection{Architecture}
\label{sec:architecture}
\vspace{-0.05in}

The formulation of UFC's control adapter in Eq.~\eqref{eq:matching} is general and can be incorporated into various diffusion backbones and adapter architectures. 
In this section, we describe the instantiation of UFC within the ControlNet framework~\cite{zhang2023adding} due to its popularity and strong performance.
For clarity, we consider the pre-trained Stable Diffusion~\cite{sdv15,sdv3} with UNet backbone~\cite{sdv15} in this section, while the one with DiT backbone~\cite{sdv3} is described in Appendix~\ref{supp:dit}.

We initialize the condition encoder $g_\tau$ and image encoder $f$ as separate copies of the UNet encoder of the pre-trained diffusion model, where the image encoder is kept frozen.
For the task-specific parameters $\theta_\tau$ of the condition encoder, we adopt bias-tuning~\cite{bias-tuning}, which has been proven to be flexible for few-shot learning in dense visual prediction~\cite{kim2023universal,kim2024chameleon}.
The matching module is implemented as a multi-head cross-attention block~\cite{vaswani2017attention}, where the timestep embedding is fused through \textit{adaptive layer normalization}~\cite{perez2018film, dit} inside the attention mechanism.
More implementation details about the matching module can be found in Appendix~\ref{supp:matching}.

To flexibly control the diffusion model, we perform patch-wise matching (Eq.~\eqref{eq:matching}) at multiple levels of the UNet encoders $g_\tau$ and $f$ using $L$ separate matching modules, where $L$ denotes the number of UNet encoder layers.
The extracted features from $g_\tau$ and $f$ at each encoder layer $l$ are passed through the $l$-th matching module to obtain the corresponding control features $\mathcal{I}^l(y_\tau; \mathcal{S}_\tau)$.
These control features pass through a zero-initialized linear projection layer $\mathcal{Z}^l$ and are added to the original features of the target diffusion model:
\begin{equation}
e^l_c = e^l + \mathcal{Z}^l(\mathcal{I}^l(y_\tau; \mathcal{S}_\tau)),
\label{eq:decoder_skip_connection}
\end{equation}
where $e^l$ is the output from the $l$-th layer of the diffusion encoder.
Finally, the resulting feature $e^l_c$ replaces the activation $e^l$ and is integrated via skip connections into the corresponding UNet decoder layer of the diffusion model.

\vspace{-0.05in}
\subsection{Training and Inference}
\label{sec:training}
\vspace{-0.05in}
To equip UFC with few-shot learning capabilities, we adopt the standard episodic meta-training protocol, wherein each training iteration is an episode that mirrors the test-time procedure. 
In each episode, we first sample a specific control task $\tau$ from a meta-training dataset $\mathcal{D}{_{\text{train}}}$ that consists of multiple tasks.
The sampled task $\tau$ supplies a support set $\mathcal{S}_\tau$ and a query set $Q_{\tau}$, where the model exploits the support set to control the diffusion model with the query condition.
The model is optimized using the standard denoising loss~\cite{ho2020denoising} (or its flow-matching variant~\cite{lipman2023flow}):
\begin{equation}
    \displaystyle \min_{\theta, \theta_\tau, \sigma, \mathcal{Z}} 
    \mathbb{E}_{\mathcal{S}_\tau,Q_\tau \sim \mathcal{D}{_\text{train}}} \mathbb{E}_{ (z_0, y_\tau, c_\text{text})\sim Q_\tau, t,\epsilon} \left[ \|\epsilon - \mathcal{E}_\phi(z_t, c_{\text{text}}, t, \mathcal{I}(y_{\tau}^q; \mathcal{S}_\tau))\|^2 \right]. 
\end{equation}

\vspace{-0.05in}
During meta-training, the parameters $\theta$ and $\theta_\tau$ in the condition encoder, the matching modules $\sigma$, and the projection layers $\mathcal{Z}$ are updated while the pre-trained diffusion model $\mathcal{E}_\phi$ and the image encoder $f$ remain frozen.
By repeatedly training on the few-shot learning episodes, the model learns a generalizable mapping from the query condition and the support set to the unified control features.

For an unseen task $\tau_\text{novel}$, we perform fine-tuning using its support set $\mathcal{S}_{\tau_\text{novel}}$.
To this end, we randomly partition the support set into two disjoint subsets $\mathcal{S}_{\tau_{\text{novel}}} = \tilde{\mathcal{S}} \cup \tilde{\mathcal{Q}}$, where $\tilde{\mathcal{S}}$ and $\tilde{\mathcal{Q}}$ act as the pseudo-support and the query sets, respectively.
Then the universal control adapter is fine-tuned with an objective similar to the meta-training:
\vspace{-0.06in}
\begin{equation}
    \displaystyle \min_{\theta_\tau, \sigma, \mathcal{Z}} 
    \mathbb{E}_{\tilde{\mathcal{S}},\tilde{\mathcal{Q}} \sim \mathcal{S}_{\tau_\text{novel}}} \mathbb{E}_{ (z_0, y_\tau, c_\text{text})\sim \tilde{\mathcal{Q}}, t, \epsilon} \left[ \|\epsilon - \mathcal{E}_\phi(z_t, c_{\text{text}}, t, \mathcal{I}(y_{\tau}; \tilde{\mathcal{S}}))\|^2 \right].
\end{equation}
Fine-tuning a small number of parameters $\theta_\tau, \sigma, \mathcal{Z}$, while freezing the shared parameters $\theta$, makes the model robust to over-fitting to the few-shot support set.
After fine-tuning, UFC can produce a control feature for unseen condition types $\tau_\text{novel}$ using the support set $\mathcal{S}_{\tau_{\text{novel}}}$ as described in Section~\ref{sec:adapter}.
\vspace{-0.06in}
\section{Experiments}
\vspace{-0.06in}

\subsection{Experimental setup}
\label{sec:exp_setup}
\vspace{-0.06in}

\paragraph{Model settings} For experiments using the UNet architecture, we adopt Stable Diffusion v1.5~\cite{sdv15} as the diffusion backbone.
Spatial conditions and support images are first projected into the latent space using the pre-trained VAE~\cite{vae} from the diffusion model, then processed by respective encoders. When encoding these inputs, we fix the denoising timestep to zero, provide empty text prompts, and omit the noisy latent input in order to focus solely on spatial information. Detailed model setting using DiT backbone is provided in the Appendix~\ref{supp:dit}.

\vspace{-0.06in}
\paragraph{Datasets} To enable episodic meta-training, we sample 300K text-image pairs from LAION-400M~\cite{schuhmann2021laion}, with 150K containing humans (for Pose and Densepose) and 150K randomly sampled. Images are resized to the resolution $512 \times 512$. Spatial conditions are extracted using pretrained off-the-shelf models. We consider six condition modalities that cover diverse semantics: Canny edge~\cite{Canny}, HED edge~\cite{hed}, MiDaS Depth and surface Normal~\cite{midas}, human Pose~\cite{openpose}, and human Densepose~\cite{guler2018densepose}. We consider Densepose as a control signal for semantic segmentation to constrain all segmentation classes to be included in the support set. During the training process, we maintain the setting that the models will be trained on 150K image-condition pairs for each task. 

To evaluate the performance over novel conditions, the six tasks are grouped into three disjoint splits: (Canny, HED), (Normal, Depth), and (Pose, Densepose).
Models are trained on two splits and evaluated on the remaining one. These splits ensure test tasks differ significantly from training tasks, providing a robust measure of few-shot generalization. 
To simulate few-shot learning, we randomly sample support sets and evaluate the performance on the remaining ones, with the exception of human pose tasks (Pose and DensePose), for which we manually select support examples to ensure coverage of diverse human poses.

\vspace{-0.06in}
\paragraph{Implementation details} 
We train UFC (UNet diffusion backbone) for 12.5K iterations with a batch size of 96 on 8 NVIDIA RTX 3090 GPUs, using AdamW~\cite{adamw} with a learning rate of $1 \times 10^{-5}$.
During inference with the UNet backbone, we adopt the PNDM sampler~\cite{pndm} with 50 denoising steps, classifier-free guidance (CFG) \cite{ho2021classifierfree} scale of 7.5, and seed 42. 
For testing, we fine-tune UFC using all image–condition pairs in the support set, but use only a subset of them as conditions during inference to accommodate memory constraints and reduce generation-time overhead.

\vspace{-0.06in}
\paragraph{Baselines}  We compare our method in the main experiment against three types of baselines:
\vspace{-0.06in}
\begin{itemize}[leftmargin=*]
    \item \textbf{Fully-supervised baselines} ControlNet~\cite{zhang2023adding} and Uni-ControlNet~\cite{zhao2023uni} are both trained in a fully-supervised manner on the entire dataset (150K images for each task). ControlNet is trained separately for each condition type, while Uni-ControlNet is jointly trained with all conditions.
    \item \textbf{Few-shot baselines} As the first to tackle spatial control in diffusion models under a few-shot setting, we adapt existing methods, Uni-ControlNet and Prompt Diffusion~\cite{wang2023incontext}, to align with our few-shot adaptation. For Uni-ControlNet, as mentioned in the original paper, we extend the input layer with additional channels for novel conditions and few-shot fine-tune this layer. For Prompt Diffusion, we evaluate both the zero-shot setup as proposed in the original paper and the full fine-tuning setup. 
    For fair comparison, these few-shot methods are pre-trained using the same meta-training dataset as our method and fine-tuned on the same few-shot data.
    \item \textbf{Training-free baselines} 
    We also compare our method with training-free methods, Ctrl-X~\cite{ctrl_x} and FreeControl~\cite{mo2024freecontrol}, which directly employ the frozen T2I diffusion models for spatial control. 
    We consider Ctrl-X as our main evaluation but leave comparisons with FreeControl in Appendix~\ref{supp:baselines} using a smaller evaluation set due to its slow inference speed.
\end{itemize}

\vspace{-0.1in}
\paragraph{Evaluation Protocol} 
We quantitatively evaluate all methods on COCO 2017~\cite{coco2017} validation split, which contains 5,000 images with associated captions. All images and their corresponding conditions are resized to $512 \times 512$. For tasks Pose and Densepose, we restrict evaluation to images containing humans.
To assess image generation quality, we report the Fréchet Inception Distance (FID)~\cite{fid}. To evaluate controllability, we extract spatial conditions from the generated images using the same pre-trained networks used for dataset construction and compare them with the query conditions using task-specific metrics. Specifically, we use the Structural Similarity Index (SSIM) for Canny and HED, Average Precision ($\text{AP}^{50}$) computed using Object Keypoint Similarity for human Pose, Mean Squared Error (MSE) for Depth prediction, Mean Angular Error (MAE) for surface Normal, and mean Intersection over Union (mIoU) for human segmentation in Densepose.

\subsection{Main Results}

\begin{table}[t]
\caption{Controllability measurement on COCO 2017. Few-shot baselines are fine-tuned with \textbf{30 shots} for evaluation on novel tasks.}
\label{main_table_controllability}
\centering
\footnotesize
\resizebox{\linewidth}{!}{
\setlength{\tabcolsep}{4pt}
\begin{tabular}{c|c| c c c c c c}
\multirow{2}{*}{Baseline Type} & \multirow{2}{*}{Method}& Canny & HED & Depth & Normal & Pose & Densepose \\
& & SSIM ($\uparrow$) & SSIM ($\uparrow$) & MSE ($\downarrow$) & MAE ($\downarrow$) & $\text{AP}^{50}$ ($\uparrow$) & mIoU ($\uparrow$) \\ \midrule

\multirow{2}{*}{Fully-supervised} & ControlNet~\cite{zhang2023adding} & \textbf{0.3598} & \textbf{0.5972} & \textbf{89.09} & \textbf{14.14} & \textbf{0.525} & \textbf{0.4824} \\
& Uni-ControlNet~\cite{zhao2023uni} & 0.3378 & 0.5808 & 92.60 & 14.59 & 0.340 & 0.4695 \\ \midrule

Training-free & Ctrl-X~\cite{ctrl_x} & 0.2901 & 0.3002 & 98.10 & 19.38 & 0.005 & 0.1352 \\\midrule

\multirow{4}{*}{Few-shot} &
Prompt Diffusion~\cite{wang2023incontext} & 0.2120 & 0.2887 & 98.81 & 20.34 & 0.012 & 0.2266 \\ 
& Uni-ControlNet + FT~\cite{zhao2023uni} & 0.2222 & 0.2917 & 99.21 & 20.36 & 0.010 & 0.2687 \\
& Prompt Diffusion + FT~\cite{wang2023incontext} & 0.2773 & 0.4810 & 95.64 & 18.04 & 0.034 & 0.3548 \\
& \textbf{UFC (Ours - UNet)} & \textbf{0.3239} & \textbf{0.5121} & \textbf{94.38} & \textbf{15.09} & \textbf{0.229} & \textbf{0.4340}\\ \bottomrule

\end{tabular}
}
\end{table}




\begin{table}[t]
\caption{FID evaluation on COCO 2017. Few-shot baselines are fine-tuned with \textbf{30 shots} for evaluation on novel tasks.}
\label{main_table_fid}
\centering
\footnotesize
\resizebox{0.95\linewidth}{!}{
\begin{tabular}{c|c| c c c c c c}
Baseline Type & Method & Canny & HED & Depth & Normal & Pose & Densepose \\ \midrule

\multirow{2}{*}{Fully-supervised} & ControlNet~\cite{zhang2023adding} & 21.06 & 17.15 & \textbf{20.16} & 21.73 & \textbf{44.33} & \textbf{34.72} \\
& Uni-ControlNet~\cite{zhao2023uni} & \textbf{16.79} & \textbf{16.76} & 20.35 & \textbf{19.59} & 47.13 & 36.77 \\\midrule

Training-free & Ctrl-X~\cite{wang2023incontext} & 29.83 & 30.18 & 30.32 & 30.35 & 56.60 & 45.88 \\\midrule

\multirow{4}{*}{Few-shot} &
Prompt Diffusion~\cite{wang2023incontext} & 24.85 & 23.00 & 24.07 & 23.28 & \textbf{43.60} & \textbf{34.15} \\
& Uni-ControlNet~\cite{zhao2023uni} + FT & 24.58 & 25.48 & 24.77 & 25.12 & 47.82 & 36.06 \\
& Prompt Diffusion~\cite{wang2023incontext}  + FT & 20.57 & \textbf{19.86} & 24.56 & 24.30 & 45.94 & 36.83 \\
& \textbf{UFC (Ours - UNet)} & \textbf{19.24} & 20.58 & \textbf{21.04} & \textbf{21.60} & 47.91 & 37.79 \\ \bottomrule

\end{tabular}
}
\vspace{-0.05in}
\end{table}

\begin{figure}[t]
  \centering
  \vspace{0.5cm}
  \includegraphics[width=1\linewidth]{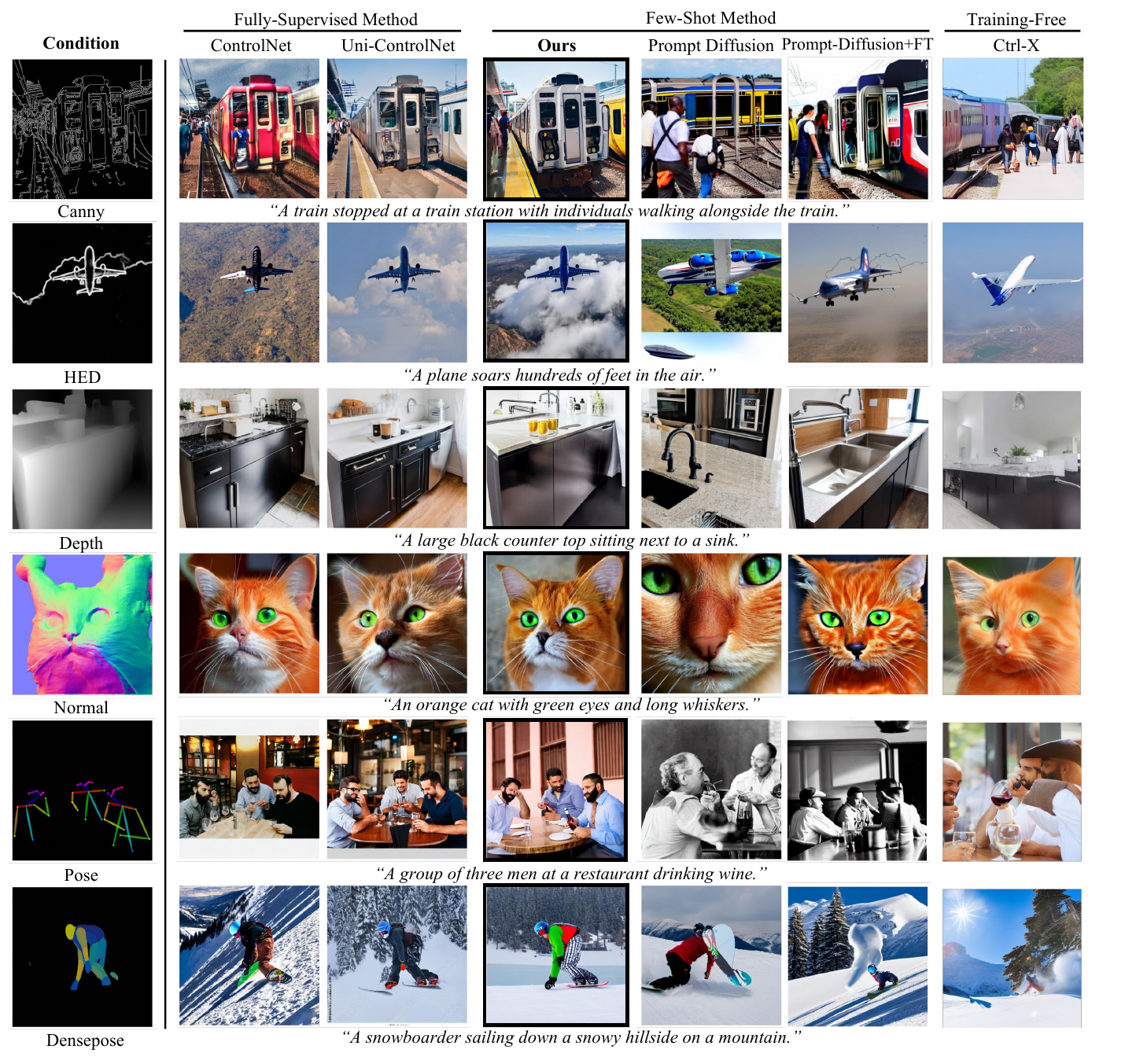}
  \vspace{-0.2in}
  \caption{
    Qualitative comparison across six spatial control tasks. Our method (highlighted in black boxes), fine-tuned with \textbf{30-shot} on unseen tasks, demonstrates competitive controllability with fully supervised baselines. In contrast, other baselines struggle to follow the spatial guidance accurately. 
    }
    \label{fig:main}
\end{figure}

\paragraph{Quantitative Results} 
We report the \textbf{30-shot} performance of our model with UNet backbone~\cite{sdv15} in terms of two aspects: controllability (Table~\ref{main_table_controllability}) and image quality (Table~\ref{main_table_fid}). Despite using only a few annotated examples, UFC achieves comparable image quality and over 90\% of the controllability performance of fully supervised baselines across most conditions. We observe that dense condition maps benefit more from the matching mechanism, as fine-grained similarity modeling enables more effective control features. This is reflected in the smaller relative performance gap with fully supervised baselines on Densepose compared to Pose.

UFC consistently outperforms all few-shot and training-free methods in controllability while maintaining equal or better image quality. 
Among the few-shot baselines, Uni-ControlNet shows clear underfitting, and zero-shot Prompt Diffusion fails to generalize, both having weak controllability and FID. 
Compared to fine-tuned (FT) Prompt Diffusion, UFC has significant gains in both metrics on Canny, Depth, and Normal, with up to 16.8\% improvement in controllability. For Densepose and HED, our method achieves similar FID scores but significantly exceeds FT Prompt Diffusion in controllability. On Pose, although FT Prompt Diffusion yields lower FID, its structural control is ineffective ($\text{AP}^{50}\approx 0$), while UFC achieves 0.229.

UFC’s strong few-shot performance stems from two factors. 
First, it effectively handles the large discrepancy between condition modalities by adapting the interpolation of support image features into task-specific control signals, unlike Prompt Diffusion or Uni-ControlNet, which lack robust strategies for handling distribution shifts.
Second, UFC introduces an efficient adaptation mechanism by fine-tuning the matching module $\sigma$, projection layers $\mathcal{Z}$, and task-specific parameters $\theta_\tau$ in the condition encoder from the meta-training weights. In contrast, Uni-ControlNet updates only a tiny subset of parameters, leading to underfitting, while Prompt Diffusion lacks any explicit adaptation mechanism, and naively full fine-tuning is ineffective in the few-shot regime.

\paragraph{Qualitative Results} 
We present qualitative comparisons in Figure~\ref{fig:main}. As shown, our method with UNet backbone, fine-tuned with only \textbf{30 shots}, consistently adheres to the given query conditions across various novel tasks. 
In contrast, the images generated by the training-free method, Ctrl-X, exhibit low generation quality and poor alignment to the spatial condition, indicating its limited adaptability to diverse spatial conditions. 
Similarly, zero-shot Prompt Diffusion fails to follow spatial guidance, suggesting that its in-context learning mechanism struggles when there are significant differences between training and unseen tasks. 
Lastly, while fine-tuned Prompt Diffusion loosely follows the spatial inputs, it often introduces undesired visual artifacts, particularly under the HED and DensePose conditions.    

\subsection{Analysis}
\label{sec:analysis}

\begin{table}[t]
\caption{Ablation study on matching module and parameters adaptation.}
\label{table_ablation}
\footnotesize
\centering
\begingroup            
\setlength{\tabcolsep}{3pt}   
\begin{tabular}{@{}l*{12}{c}@{}}   
\toprule
\multirow{2}{*}{Methods} & \multicolumn{2}{c}{Canny} & \multicolumn{2}{c}{HED}
                         & \multicolumn{2}{c}{Depth} & \multicolumn{2}{c}{Normal}
                         & \multicolumn{2}{c}{Pose} & \multicolumn{2}{c}{Densepose} \\
\cmidrule(l){2-3} \cmidrule(l){4-5} \cmidrule(l){6-7}
\cmidrule(l){8-9} \cmidrule(l){10-11} \cmidrule(l){12-13}
 & SSIM$\uparrow$ & FID & SSIM$\uparrow$ & FID
 & MSE$\downarrow$ & FID & MAE$\downarrow$ & FID
 & $\text{AP}^{50}$$\uparrow$ & FID & mIoU$\uparrow$ & FID \\ \midrule
\textbf{UFC (UNet) }               & \textbf{0.3239} & \textbf{19.24} & \textbf{0.5121} & 20.58
                    & \textbf{94.38}  & \textbf{21.04} & \textbf{15.09}  & \textbf{21.60}
                    & \textbf{0.229}  & \textbf{47.91} & \textbf{0.4340} & 37.79 \\
w\textbackslash o Matching
                    & 0.2984 & 20.43 & 0.4972 & \textbf{20.43}
                    & 96.17  & 21.10 & 16.06  & 23.70
                    & 0.150  & 48.47 & 0.3995 & 40.84 \\
w\textbackslash o Fine-tuning
                    & 0.2443 & 21.64 & 0.3688 & 26.42
                    & 97.92  & 22.12 & 18.90  & 22.82
                    & 0.002  & 47.97 & 0.1855 & \textbf{35.59} \\ \bottomrule
\end{tabular}
\endgroup          
\end{table}
\paragraph{Ablation study} 
To assess the effectiveness of our design in handling the large discrepancy across condition modalities, we conduct an ablation study with two variants of UFC on UNet backbone~\cite{sdv15}: (1) \textbf{UFC w/o Matching} and (2) \textbf{UFC w/o Fine-tuning}. Both variants are trained on the same dataset as our main method and evaluated on the COCO 2017 validation split. 
In (1), we remove the matching mechanism and encode the query condition directly using the condition encoder \textit{i.e.}, $\mathcal{G}_\tau(y_\tau)$ while retaining task-specific parameters and meta-learning paradigm.
At test time, we fine-tune $\theta_\tau$ and $\mathcal{Z}$ using the small support set. 
In (2), we disable adaptation entirely by sharing all parameters across tasks during training and evaluating directly on unseen tasks without any fine-tuning.

Quantitative results are presented in Table~\ref{table_ablation}, with qualitative examples in Appendix~\ref{supp:shots}. 
Both ablation variants show a clear drop in performance compared to our full model. Notably, \textbf{UFC w/o Matching} achieves similar image quality but significantly lower controllability, underscoring the importance of leveraging task-agnostic features from support images via the matching mechanism. 
We provide the attention map visualization in Figure~\ref{supp:fig:attn_map}, Appendix~\ref{supp:shots} for a better understanding of our matching mechanism.
\textbf{UFC w/o Fine-tuning} performs the worst overall, highlighting the essential role of parameter adaptation in generalizing to novel spatial conditions.

\begin{figure}[!t]
  \centering
  \includegraphics[width=\linewidth]{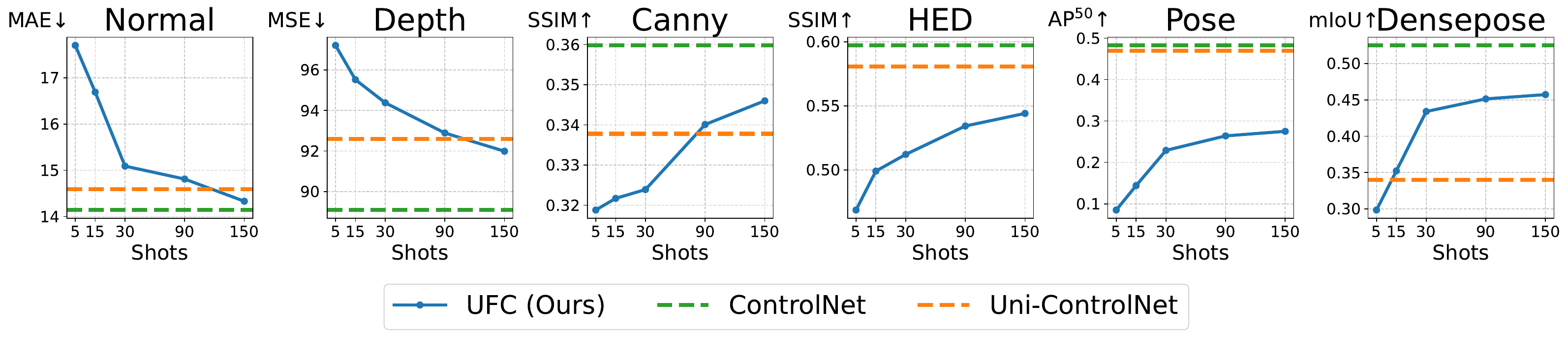}
  \vspace{-0.2in}
  \caption{
    Performance of UFC when fine-tuned with different numbers of support data. Overall, UFC consistently improves the controllability with the increasing size of support sets. The results on FID are presented in the Appendix~\ref{supp:shots}, Figure~\ref{supp:fig:fid}.
    }
    \label{fig:numshots}
\end{figure}

\paragraph{Impact of Support Set Sizes} 
We evaluate UFC on varying sizes of support sets for unseen condition types and plot the performance curves in Figure \ref{fig:numshots}. 
We can observe a clear trend that the controllability of our method improves considerably with the increasing size of the support set. 
With 150 support image-condition pairs (0.1\% of full training data), UFC begins to surpass the controllability of the fully-supervised baseline (Uni-ControlNet) on several tasks. 
We also present the FID score over varying support set sizes in Figure~\ref{supp:fig:fid} in Appendix~\ref{supp:shots}, which shows that our method maintains the image quality across different shots.
This improvement in controllability demonstrates the effectiveness of our method in adapting to unseen spatial condition modalities through lightweight fine-tuning, eliminating the costly fully-supervised training for novel tasks.

\paragraph{Impact of Support Samples} 
To evaluate the sensitivity of UFC to the choice of support set, we evaluate 30-shot performance with various support sets and report the results in Table~\ref{table_diversity}, Appendix~\ref{supp:shots}.
It shows that performance of our method tends to remain consistent under diverse support sets in most tasks.
However, in some tasks such as Pose and Densepose, we observe that the model performed worse than the main results that used curated supports maximizing diversity in scale, deformation, crowding, and occlusion. 
It shows that the importance of support-set diversity is itself task-dependent, and ensuring sample diversity can potentially benefit the few-shot performance.

\begin{table}[t]
\caption{Quantitative evaluation of UFC with different backbones in \textit{30-shot} setting.}
\label{table_dit}
\footnotesize
\centering
\begingroup            
\setlength{\tabcolsep}{3pt}   
\begin{tabular}{@{}l*{12}{c}@{}}   
\toprule
\multirow{2}{*}{Backbone} & \multicolumn{2}{c}{Canny} & \multicolumn{2}{c}{HED}
                         & \multicolumn{2}{c}{Depth} & \multicolumn{2}{c}{Normal}
                         & \multicolumn{2}{c}{Pose} & \multicolumn{2}{c}{Densepose} \\
\cmidrule(l){2-3} \cmidrule(l){4-5} \cmidrule(l){6-7}
\cmidrule(l){8-9} \cmidrule(l){10-11} \cmidrule(l){12-13}
 & SSIM$\uparrow$ & FID & SSIM$\uparrow$ & FID
 & MSE$\downarrow$ & FID & MAE$\downarrow$ & FID
 & $\text{AP}^{50}$$\uparrow$ & FID & mIoU$\uparrow$ & FID \\ \midrule
Ours (UNet)               & 0.3239& \textbf{19.24} & 0.5121 & 20.58
                    & 94.38  & \textbf{21.04} & 15.09  & \textbf{21.60}
                    & 0.229  &\textbf{ 47.91} & 0.4340 & \textbf{37.79} \\
Ours (DiT)          & \textbf{0.3984} & 20.43 & \textbf{0.6001} & \textbf{19.84}
                    & \textbf{91.51}  & 22.70 & \textbf{13.79}  & 23.06
                    & \textbf{0.336}  & 51.88 & \textbf{0.4708} & 39.27 \\ \bottomrule

\end{tabular}
\endgroup          
\end{table}
\paragraph{Impact of Diffusion Backbone}

Since UFC is applicable to various diffusion backbones, we employ the stronger pre-trained model based on DiT~\cite{sdv3} and compare it with the one based on UNet used in our main table.
The quantitative results are reported in Table~\ref{table_dit}, showing that the DiT-based model with a stronger diffusion backbone consistently outperforms the UNet-based model in controllability across all tasks. The lower image quality compared to the UNet variant, which is also observed in \cite{tan2024ominicontrol}, can be due to the mismatch between the optimized image resolution of the pre-trained diffusion backbone with the testing resolution (1024 and 512). Figure~\ref{fig:DiT_UNet_compare} in Appendix~\ref{supp:dit} shows that UFC (DiT) qualitatively achieves more fine-grained controllability than the UNet counterpart. More qualitative examples using the DiT backbone are shown in Figure~\ref{supp:fig:dit} in Appendix.

\paragraph{Evaluation on More Novel Conditions} 
To further validate the few-shot capability of our method in more challenging settings, we evaluate our method on novel spatial conditions on 3D structures, such as 3D meshes, wireframes, and point clouds extracted from iso3d dataset~\cite{ebert20253d}. 
Importantly, these control tasks involve not only novel spatial conditions but also different output image distributions from meta-training datasets, simulating a more realistic generalization scenario to unseen control tasks.

We meta-train UFC on all six tasks in our original dataset (\emph{i.e.}, subset of LAION-400M), then fine-tune the model on 30-shot support sets manually collected for each condition type. The fine-tuned model is then tested on unseen query conditions. The qualitative results in Figure~\ref{fig:supp:new_condition},  Appendix~\ref{supp:more_results} demonstrate that generated images align closely with the spatial conditions provided by 3D meshes, wireframes, and point clouds. These results confirm UFC's effectiveness when encountering novel spatial conditions. 


\section{Conclusion}
In this paper, we present Universal Few-shot Control (UFC), a unified control adapter that is capable of few-shot controlling text-to-image diffusion models with unseen spatial conditions. UFC adapts interpolated visual features from support images into task-specific control signals, guided by the analogy between query and support conditions. When evaluated on unseen control conditions, episodic meta-trained UFC demonstrated strong few-shot ability: it consistently outperformed all few-shot baselines and achieved precise spatial control over generated images. Notably, UFC can be on par with a fully supervised baseline, despite using only 0.1\% of the full data for fine-tuning. Finally, we demonstrated that our method is applicable to both recent architectures of diffusion models, including UNet and DiT backbones. 

\paragraph{Limitations and Future Research}
While UFC demonstrates strong few-shot performance in spatially-conditioned image generation with T2I diffusion models, several limitations remain for future work. 
First, our framework is designed primarily for spatial control generation rather than tasks that require preserving the appearance of the condition image, such as style transfer, inverse problems (\textit{e.g.,} colorization, deblurring, inpainting). Extending the framework to handle such tasks can be a future research direction. 
Next, our approach requires fine-tuning on a small annotated set for each new task, unlike Large Language Models, which adapt to new tasks via in-context learning from just a few examples without fine-tuning. Developing similar capabilities for spatial control image generation remains an open and promising challenge.


\paragraph{Acknowledgments}
 This work was in part supported by the National Research Foundation of Korea (RS-2024-00351212 and RS-2024-00436165) and the Institute of Information \& communications Technology Planning \& Evaluation (IITP) (RS-2022-II220926, RS-2024-00509279, RS-2021-II212068, RS-2022-II220959, and RS-2019-II190075) funded by the Korea government (MSIT).

\newpage
\bibliography{main_arxiv}

@String(CVPR= {IEEE Conf. Comput. Vis. Pattern Recog.})

@String(ICCV= {Int. Conf. Comput. Vis.})

@String(ECCV= {Eur. Conf. Comput. Vis.})

@String(NIPS= {Adv. Neural Inform. Process. Syst.})

@String(TOG= {ACM Trans. Graph.})

@String(ICLR = {Int. Conf. Learn. Represent.})

@String(AAAI = {AAAI})

@String(CVPR  = {CVPR})

@String(ICCV  = {ICCV})

@String(ECCV  = {ECCV})

@String(NIPS  = {NeurIPS})

@String(TOG   = {ACM TOG})

@String(ICLR  = {ICLR})

@inproceedings{zhang2023adding,
  title={Adding conditional control to text-to-image diffusion models},
  author={Zhang, Lvmin and Rao, Anyi and Agrawala, Maneesh},
  booktitle={Proceedings of the IEEE/CVF international conference on computer vision},
  pages={3836--3847},
  year={2023}
}

@inproceedings{
    wang2023incontext,
    title={In-Context Learning Unlocked for Diffusion Models},
    author={Zhendong Wang and Yifan Jiang and Yadong Lu and yelong shen and Pengcheng He and Weizhu Chen and Zhangyang Wang and Mingyuan Zhou},
    booktitle={Thirty-seventh Conference on Neural Information Processing Systems},
    year={2023},
    url={https://openreview.net/forum?id=6BZS2EAkns}
}

@article{zhao2023uni,
  title={Uni-controlnet: All-in-one control to text-to-image diffusion models},
  author={Zhao, Shihao and Chen, Dongdong and Chen, Yen-Chun and Bao, Jianmin and Hao, Shaozhe and Yuan, Lu and Wong, Kwan-Yee K},
  journal={Advances in Neural Information Processing Systems},
  volume={36},
  pages={11127--11150},
  year={2023}
}

@article{tan2024ominicontrol,
  title={Ominicontrol: Minimal and universal control for diffusion transformer},
  author={Tan, Zhenxiong and Liu, Songhua and Yang, Xingyi and Xue, Qiaochu and Wang, Xinchao},
  journal={arXiv preprint arXiv:2411.15098},
  year={2024}
}

@inproceedings{
kim2023universal,
title={Universal Few-shot Learning of Dense Prediction Tasks with Visual Token Matching},
author={Donggyun Kim and Jinwoo Kim and Seongwoong Cho and Chong Luo and Seunghoon Hong},
booktitle={The Eleventh International Conference on Learning Representations },
year={2023},
url={https://openreview.net/forum?id=88nT0j5jAn}
}

@article{schuhmann2021laion,
  title={Laion-400m: Open dataset of clip-filtered 400 million image-text pairs},
  author={Schuhmann, Christoph and Vencu, Richard and Beaumont, Romain and Kaczmarczyk, Robert and Mullis, Clayton and Katta, Aarush and Coombes, Theo and Jitsev, Jenia and Komatsuzaki, Aran},
  journal={arXiv preprint arXiv:2111.02114},
  year={2021}
}

@inproceedings{sdv15,
  author       = {Robin Rombach and
                  Andreas Blattmann and
                  Dominik Lorenz and
                  Patrick Esser and
                  Bj{\"{o}}rn Ommer},
  title        = {High-Resolution Image Synthesis with Latent Diffusion Models},
  booktitle    = {{IEEE/CVF} Conference on Computer Vision and Pattern Recognition,
                  {CVPR} 2022, New Orleans, LA, USA, June 18-24, 2022},
  pages        = {10674--10685},
  publisher    = {{IEEE}},
  year         = {2022},
  url          = {https://doi.org/10.1109/CVPR52688.2022.01042},
  doi          = {10.1109/CVPR52688.2022.01042},
  bibsource    = {dblp computer science bibliography, https://dblp.org}
}

@inproceedings{sdv3,
  title={Scaling rectified flow transformers for high-resolution image synthesis},
  author={Esser, Patrick and Kulal, Sumith and Blattmann, Andreas and Entezari, Rahim and M{\"u}ller, Jonas and Saini, Harry and Levi, Yam and Lorenz, Dominik and Sauer, Axel and Boesel, Frederic and others},
  booktitle={Forty-first international conference on machine learning},
  year={2024}
}

@inproceedings{vae,
  author       = {Diederik P. Kingma and
                  Max Welling},
  editor       = {Yoshua Bengio and
                  Yann LeCun},
  title        = {Auto-Encoding Variational Bayes},
  booktitle    = {2nd International Conference on Learning Representations, {ICLR} 2014,
                  Banff, AB, Canada, April 14-16, 2014, Conference Track Proceedings},
  year         = {2014},
  url          = {http://arxiv.org/abs/1312.6114},
  bibsource    = {dblp computer science bibliography, https://dblp.org}
}

@inproceedings{bias-tuning,
    title = "{B}it{F}it: Simple Parameter-efficient Fine-tuning for Transformer-based Masked Language-models",
    author = "Ben Zaken, Elad  and
      Goldberg, Yoav  and
      Ravfogel, Shauli",
    editor = "Muresan, Smaranda  and
      Nakov, Preslav  and
      Villavicencio, Aline",
    booktitle = "Proceedings of the 60th Annual Meeting of the Association for Computational Linguistics (Volume 2: Short Papers)",
    month = may,
    year = "2022",
    address = "Dublin, Ireland",
    publisher = "Association for Computational Linguistics",
    url = "https://aclanthology.org/2022.acl-short.1/",
    doi = "10.18653/v1/2022.acl-short.1",
    pages = "1--9"
}

@inproceedings{transformer,
author = {Vaswani, Ashish and Shazeer, Noam and Parmar, Niki and Uszkoreit, Jakob and Jones, Llion and Gomez, Aidan N. and Kaiser, \L{}ukasz and Polosukhin, Illia},
title = {Attention is all you need},
year = {2017},
isbn = {9781510860964},
publisher = {Curran Associates Inc.},
address = {Red Hook, NY, USA},
booktitle = {Proceedings of the 31st International Conference on Neural Information Processing Systems},
pages = {6000–6010},
numpages = {11},
location = {Long Beach, California, USA},
series = {NIPS'17}
}

@inproceedings{unicontrol,
author = {Qin, Can and Zhang, Shu and Yu, Ning and Feng, Yihao and Yang, Xinyi and Zhou, Yingbo and Wang, Huan and Niebles, Juan Carlos and Xiong, Caiming and Savarese, Silvio and Ermon, Stefano and Fu, Yun and Xu, Ran},
title = {UniControl: a unified diffusion model for controllable visual generation in the wild},
year = {2023},
publisher = {Curran Associates Inc.},
address = {Red Hook, NY, USA},
booktitle = {Proceedings of the 37th International Conference on Neural Information Processing Systems},
articleno = {1862},
numpages = {32},
location = {New Orleans, LA, USA},
series = {NIPS '23}
}

@inproceedings{controlnetpp,
author = {Li, Ming and Yang, Taojiannan and Kuang, Huafeng and Wu, Jie and Wang, Zhaoning and Xiao, Xuefeng and Chen, Chen},
title = {ControlNet++: Improving Conditional Controls with Efficient Consistency Feedback},
year = {2024},
isbn = {978-3-031-72666-8},
publisher = {Springer-Verlag},
address = {Berlin, Heidelberg},
url = {https://doi.org/10.1007/978-3-031-72667-5_8},
doi = {10.1007/978-3-031-72667-5_8},
booktitle = {Computer Vision – ECCV 2024: 18th European Conference, Milan, Italy, September 29–October 4, 2024, Proceedings, Part VII},
pages = {129–147},
numpages = {19},
location = {Milan, Italy}
}

@ARTICLE{Canny,
  author={Canny, John},
  journal={IEEE Transactions on Pattern Analysis and Machine Intelligence}, 
  title={A Computational Approach to Edge Detection}, 
  year={1986},
  volume={PAMI-8},
  number={6},
  pages={679-698},
  doi={10.1109/TPAMI.1986.4767851}}

@INPROCEEDINGS{hed,
  author={Xie, Saining and Tu, Zhuowen},
  booktitle={2015 IEEE International Conference on Computer Vision (ICCV)}, 
  title={Holistically-Nested Edge Detection}, 
  year={2015},
  volume={},
  number={},
  pages={1395-1403},
  doi={10.1109/ICCV.2015.164}}

@InProceedings{openpose,
author = {Cao, Zhe and Simon, Tomas and Wei, Shih-En and Sheikh, Yaser},
title = {Realtime Multi-Person 2D Pose Estimation Using Part Affinity Fields},
booktitle = {Proceedings of the IEEE Conference on Computer Vision and Pattern Recognition (CVPR)},
month = {July},
year = {2017}
}

@inproceedings{t2iadapter,
  title={T2i-adapter: Learning adapters to dig out more controllable ability for text-to-image diffusion models},
  author={Mou, Chong and Wang, Xintao and Xie, Liangbin and Wu, Yanze and Zhang, Jian and Qi, Zhongang and Shan, Ying},
  booktitle={Proceedings of the AAAI conference on artificial intelligence},
  year={2024}
}

@InProceedings{ugd,
    author    = {Bansal, Arpit and Chu, Hong-Min and Schwarzschild, Avi and Sengupta, Soumyadip and Goldblum, Micah and Geiping, Jonas and Goldstein, Tom},
    title     = {Universal Guidance for Diffusion Models},
    booktitle = {Proceedings of the IEEE/CVF Conference on Computer Vision and Pattern Recognition (CVPR) Workshops},
    month     = {June},
    year      = {2023},
    pages     = {843-852}
}

@inproceedings{mo2024freecontrol,
  title={Freecontrol: Training-free spatial control of any text-to-image diffusion model with any condition},
  author={Mo, Sicheng and Mu, Fangzhou and Lin, Kuan Heng and Liu, Yanli and Guan, Bochen and Li, Yin and Zhou, Bolei},
  booktitle={Proceedings of the IEEE/CVF Conference on Computer Vision and Pattern Recognition},
  pages={7465--7475},
  year={2024}
}

@inproceedings{yu2023freedom,
  title={Freedom: Training-free energy-guided conditional diffusion model},
  author={Yu, Jiwen and Wang, Yinhuai and Zhao, Chen and Ghanem, Bernard and Zhang, Jian},
  booktitle={Proceedings of the IEEE/CVF International Conference on Computer Vision},
  pages={23174--23184},
  year={2023}
}

@inproceedings{li2023gligen,
  title={Gligen: Open-set grounded text-to-image generation},
  author={Li, Yuheng and Liu, Haotian and Wu, Qingyang and Mu, Fangzhou and Yang, Jianwei and Gao, Jianfeng and Li, Chunyuan and Lee, Yong Jae},
  booktitle={Proceedings of the IEEE/CVF conference on computer vision and pattern recognition},
  pages={22511--22521},
  year={2023}
}

@article{midas,
  title={Towards robust monocular depth estimation: Mixing datasets for zero-shot cross-dataset transfer},
  author={Ranftl, Ren{\'e} and Lasinger, Katrin and Hafner, David and Schindler, Konrad and Koltun, Vladlen},
  journal={IEEE transactions on pattern analysis and machine intelligence},
  volume={44},
  number={3},
  pages={1623--1637},
  year={2020},
  publisher={IEEE}
}

@inproceedings{guler2018densepose,
  title={Densepose: Dense human pose estimation in the wild},
  author={G{\"u}ler, R{\i}za Alp and Neverova, Natalia and Kokkinos, Iasonas},
  booktitle={Proceedings of the IEEE conference on computer vision and pattern recognition},
  pages={7297--7306},
  year={2018}
}

@inproceedings{
podell2024sdxl,
title={{SDXL}: Improving Latent Diffusion Models for High-Resolution Image Synthesis},
author={Dustin Podell and Zion English and Kyle Lacey and Andreas Blattmann and Tim Dockhorn and Jonas M{\"u}ller and Joe Penna and Robin Rombach},
booktitle={The Twelfth International Conference on Learning Representations},
year={2024},
url={https://openreview.net/forum?id=di52zR8xgf}
}

@inproceedings{
saharia2022photorealistic,
title={Photorealistic Text-to-Image Diffusion Models with Deep Language Understanding},
author={Chitwan Saharia and William Chan and Saurabh Saxena and Lala Li and Jay Whang and Emily Denton and Seyed Kamyar Seyed Ghasemipour and Raphael Gontijo-Lopes and Burcu Karagol Ayan and Tim Salimans and Jonathan Ho and David J. Fleet and Mohammad Norouzi},
booktitle={Advances in Neural Information Processing Systems},
editor={Alice H. Oh and Alekh Agarwal and Danielle Belgrave and Kyunghyun Cho},
year={2022},
url={https://openreview.net/forum?id=08Yk-n5l2Al}
}

@article{lhhuang2023composer,
  title={Composer: Creative and Controllable Image Synthesis with Composable Conditions},
  author={Huang, Lianghua and Chen, Di and Liu, Yu and Yujun, Shen and Zhao, Deli and Jingren, Zhou},
  journal={arXiv preprint arxiv:2302.09778},
  year={2023}
}

@article{zhu2022few,
  title={Few-shot image generation with diffusion models},
  author={Zhu, Jingyuan and Ma, Huimin and Chen, Jiansheng and Yuan, Jian},
  journal={arXiv preprint arXiv:2211.03264},
  year={2022}
}

@article{giannone2022few,
  title={Few-shot diffusion models},
  author={Giannone, Giorgio and Nielsen, Didrik and Winther, Ole},
  journal={arXiv preprint arXiv:2205.15463},
  year={2022}
}

@article{yang2024few,
  title={Few-shot diffusion models escape the curse of dimensionality},
  author={Yang, Ruofeng and Jiang, Bo and Chen, Cheng and Wang, Baoxiang and Li, Shuai and others},
  journal={Advances in Neural Information Processing Systems},
  volume={37},
  pages={68528--68558},
  year={2024}
}

@article{sinha2021few,
  title={D2c: Diffusion-decoding models for few-shot conditional generation},
  author={Sinha, Abhishek and Song, Jiaming and Meng, Chenlin and Ermon, Stefano},
  journal={Advances in Neural Information Processing Systems},
  volume={34},
  pages={12533--12548},
  year={2021}
}

@InProceedings{Li_2023_ICCV_few,
    author    = {Li, Lingxiao and Zhang, Yi and Wang, Shuhui},
    title     = {The Euclidean Space is Evil: Hyperbolic Attribute Editing for Few-shot Image Generation},
    booktitle = {Proceedings of the IEEE/CVF International Conference on Computer Vision (ICCV)},
    month     = {October},
    year      = {2023},
    pages     = {22714-22724}
}

@misc{jin2025dualinterrelateddiffusionmodelfewshot,
      title={Dual-Interrelated Diffusion Model for Few-Shot Anomaly Image Generation}, 
      author={Ying Jin and Jinlong Peng and Qingdong He and Teng Hu and Jiafu Wu and Hao Chen and Haoxuan Wang and Wenbing Zhu and Mingmin Chi and Jun Liu and Yabiao Wang},
      year={2025},
      eprint={2408.13509},
      archivePrefix={arXiv},
      primaryClass={cs.CV},
      url={https://arxiv.org/abs/2408.13509}, 
}

@inproceedings{cao2024few,
  title={Few-shot image generation by conditional relaxing diffusion inversion},
  author={Cao, Yu and Gong, Shaogang},
  booktitle={European Conference on Computer Vision},
  pages={20--37},
  year={2024},
  organization={Springer}
}

@article{gu2024analogist,
  title={Analogist: Out-of-the-box visual in-context learning with image diffusion model},
  author={Gu, Zheng and Yang, Shiyuan and Liao, Jing and Huo, Jing and Gao, Yang},
  journal={ACM Transactions on Graphics (TOG)},
  volume={43},
  number={4},
  pages={1--15},
  year={2024},
  publisher={ACM New York, NY, USA}
}

@article{yang2023imagebrush,
  title={Imagebrush: Learning visual in-context instructions for exemplar-based image manipulation},
  author={Yang, Yifan and Peng, Houwen and Shen, Yifei and Yang, Yuqing and Hu, Han and Qiu, Lili and Koike, Hideki and others},
  journal={Advances in Neural Information Processing Systems},
  volume={36},
  pages={48723--48743},
  year={2023}
}

@inproceedings{pndm,
    title={Pseudo Numerical Methods for Diffusion Models on Manifolds},
    author={Luping Liu and Yi Ren and Zhijie Lin and Zhou Zhao},
    booktitle={International Conference on Learning Representations},
    year={2022},
    url={https://openreview.net/forum?id=PlKWVd2yBkY}
}

@inproceedings{
adamw,
title={Decoupled Weight Decay Regularization},
author={Ilya Loshchilov and Frank Hutter},
booktitle={International Conference on Learning Representations},
year={2019},
url={https://openreview.net/forum?id=Bkg6RiCqY7},
}

@inproceedings{
ho2021classifierfree,
title={Classifier-Free Diffusion Guidance},
author={Jonathan Ho and Tim Salimans},
booktitle={NeurIPS 2021 Workshop on Deep Generative Models and Downstream Applications},
year={2021},
url={https://openreview.net/forum?id=qw8AKxfYbI}
}

@inproceedings{coco2017,
title={Microsoft coco: Common objects in context},
author={Lin, Tsung-Yi and Maire, Michael and Belongie, Serge and Hays, James and Perona, Pietro and Ramanan, Deva and Doll{'a}r, Piotr and Zitnick, C Lawrence},
booktitle={Computer Vision--ECCV 2014: 13th European Conference, Zurich, Switzerland, September 6-12, 2014, Proceedings, Part V 13},
pages={740--755},
year={2014},
organization={Springer}
}

@inproceedings{fid,
 author = {Heusel, Martin and Ramsauer, Hubert and Unterthiner, Thomas and Nessler, Bernhard and Hochreiter, Sepp},
 booktitle = {Advances in Neural Information Processing Systems},
 editor = {I. Guyon and U. Von Luxburg and S. Bengio and H. Wallach and R. Fergus and S. Vishwanathan and R. Garnett},
 pages = {},
 publisher = {Curran Associates, Inc.},
 title = {GANs Trained by a Two Time-Scale Update Rule Converge to a Local Nash Equilibrium},
 url = {https://proceedings.neurips.cc/paper_files/paper/2017/file/8a1d694707eb0fefe65871369074926d-Paper.pdf},
 volume = {30},
 year = {2017}
}

@article{ho2020denoising,
  title={Denoising diffusion probabilistic models},
  author={Ho, Jonathan and Jain, Ajay and Abbeel, Pieter},
  journal={Advances in neural information processing systems},
  volume={33},
  pages={6840--6851},
  year={2020}
}

@inproceedings{kim2024chameleon,
  title={Chameleon: A data-efficient generalist for dense visual prediction in the wild},
  author={Kim, Donggyun and Cho, Seongwoong and Kim, Semin and Luo, Chong and Hong, Seunghoon},
  booktitle={European Conference on Computer Vision},
  pages={422--441},
  year={2024},
  organization={Springer}
}

@inproceedings{tumanyan2023plug,
  title={Plug-and-play diffusion features for text-driven image-to-image translation},
  author={Tumanyan, Narek and Geyer, Michal and Bagon, Shai and Dekel, Tali},
  booktitle={Proceedings of the IEEE/CVF Conference on Computer Vision and Pattern Recognition},
  pages={1921--1930},
  year={2023}
}

@misc{stable_diffusion_v1_5,
  author       = {CompVis},
  title        = {Stable Diffusion v1-5},
  howpublished = {\url{https://huggingface.co/stable-diffusion-v1-5/stable-diffusion-v1-5}},
  note         = {Accessed: 2025-05-22},
  year = {2022}
}

@misc{Jocher_Ultralytics_YOLO_2023,
author = {Jocher, Glenn and Qiu, Jing and Chaurasia, Ayush},
license = {AGPL-3.0},
month = jan,
title = {{Ultralytics YOLO}},
url = {https://github.com/ultralytics/ultralytics},
version = {8.0.0},
year = {2023}
}

@inproceedings{perez2018film,
  title={Film: Visual reasoning with a general conditioning layer},
  author={Perez, Ethan and Strub, Florian and De Vries, Harm and Dumoulin, Vincent and Courville, Aaron},
  booktitle={Proceedings of the AAAI conference on artificial intelligence},
  year={2018}
}

@article{inceptionscore,
  title={Improved techniques for training gans},
  author={Salimans, Tim and Goodfellow, Ian and Zaremba, Wojciech and Cheung, Vicki and Radford, Alec and Chen, Xi},
  journal={Advances in neural information processing systems},
  volume={29},
  year={2016}
}

@inproceedings{dit,
  title={Scalable diffusion models with transformers},
  author={Peebles, William and Xie, Saining},
  booktitle={Proceedings of the IEEE/CVF international conference on computer vision},
  pages={4195--4205},
  year={2023}
}

@misc{stablediffusion3.5medium,
  author       = {Stability AI},
  title        = {Stable Diffusion 3.5 Medium},
  year         = {2024},
  howpublished = {\url{https://huggingface.co/stabilityai/stable-diffusion-3.5-medium}},
  note         = {Accessed: 2025-05-22}
}

@misc{flux2024,
    author={Black Forest Labs},
    title={FLUX},
    year={2024},
    howpublished={\url{https://github.com/black-forest-labs/flux}},
}

@inproceedings{lipman2023flow,
    title={Flow Matching for Generative Modeling},
    author={Yaron Lipman and Ricky T. Q. Chen and Heli Ben-Hamu and Maximilian Nickel and Matthew Le},
    booktitle={The Eleventh International Conference on Learning Representations },
    year={2023},
    url={https://openreview.net/forum?id=PqvMRDCJT9t}
}

@article{wang2025unicombine,
  title={UniCombine: Unified Multi-Conditional Combination with Diffusion Transformer},
  author={Wang, Haoxuan and Peng, Jinlong and He, Qingdong and Yang, Hao and Jin, Ying and Wu, Jiafu and Hu, Xiaobin and Pan, Yanjie and Gan, Zhenye and Chi, Mingmin and others},
  journal={CoRR},
  year={2025}
}

@inproceedings{chen2024pixart,
  title={PixArt-$\alpha$: Fast Training of Diffusion Transformer for Photorealistic Text-to-Image Synthesis},
  author={Chen, Junsong and Yu, Jincheng and Ge, Chongjian and Yao, Lewei and Xie, Enze and Wang, Zhongdao and Kwok, James T and Luo, Ping and Lu, Huchuan and Li, Zhenguo},
  booktitle={ICLR},
  year={2024}
}

@inproceedings{chen2024pixartdelta,
  title={PIXART-$\delta$: Fast and Controllable Image Generation with Latent Consistency Models},
  author={Chen, Junsong and Luo, Simian and Xie, Enze},
  booktitle={ICML 2024 Workshop on Theoretical Foundations of Foundation Models},
  year={2024}
}

@inproceedings{sohl2015deep,
  title={Deep unsupervised learning using nonequilibrium thermodynamics},
  author={Sohl-Dickstein, Jascha and Weiss, Eric and Maheswaranathan, Niru and Ganguli, Surya},
  booktitle={International conference on machine learning},
  pages={2256--2265},
  year={2015},
  organization={pmlr}
}

@inproceedings{unet,
  title={U-net: Convolutional networks for biomedical image segmentation},
  author={Ronneberger, Olaf and Fischer, Philipp and Brox, Thomas},
  booktitle={Medical image computing and computer-assisted intervention--MICCAI 2015: 18th international conference, Munich, Germany, October 5-9, 2015, proceedings, part III 18},
  pages={234--241},
  year={2015},
  organization={Springer}
}

@article{vaswani2017attention,
  title={Attention is all you need},
  author={Vaswani, Ashish and Shazeer, Noam and Parmar, Niki and Uszkoreit, Jakob and Jones, Llion and Gomez, Aidan N and Kaiser, {\L}ukasz and Polosukhin, Illia},
  journal={Advances in neural information processing systems},
  volume={30},
  year={2017}
}

@article{lora,
  title={Lora: Low-rank adaptation of large language models.},
  author={Hu, Edward J and Shen, Yelong and Wallis, Phillip and Allen-Zhu, Zeyuan and Li, Yuanzhi and Wang, Shean and Wang, Lu and Chen, Weizhu and others},
  journal={ICLR},
  volume={1},
  number={2},
  pages={3},
  year={2022}
}

@article{ctrl_x,
  title={Ctrl-x: Controlling structure and appearance for text-to-image generation without guidance},
  author={Lin, Kuan Heng and Mo, Sicheng and Klingher, Ben and Mu, Fangzhou and Zhou, Bolei},
  journal={Advances in Neural Information Processing Systems},
  volume={37},
  pages={128911--128939},
  year={2024}
}

@article{ebert20253d,
  title={3D Arena: An Open Platform for Generative 3D Evaluation},
  author={Ebert, Dylan},
  journal={arXiv preprint arXiv:2506.18787},
  year={2025}
}
\bibliographystyle{acl_natbib}


\clearpage
\appendix

\section{Implementation Details}
\label{supp:implementation_details}

\subsection{Implementation Details on Matching Module}
\label{supp:matching}
We implement our matching module described in Section~\ref{sec:architecture} using the multi-head attention mechanism \cite{transformer}. To incorporate denoising timestep $t$, we adopt \textit{adaptive normalization} \cite{perez2018film}, modulating the output of the matching module using the denoising timestep. Specifically, we use the embedding of the query condition \( g_{\tau}(y_\tau) \), the support conditions \( \{g_{\tau}(y^i_\tau)\}_{i=1}^N \), and the support images \( \{f(x^i)\}_{i=1}^N \) as the query, key, and value inputs, respectively. Let \( Q \in \mathbb{R}^{M \times d} \), \( K, V \in \mathbb{R}^{(N \cdot M) \times d} \), and the timestep embedding be denoted as \( t_{\text{emb}} \in \mathbb{R}^{d_t} \), our matching mechanism operates as below:
\begin{equation}
    Q = \text{LayerNorm}(Q), \quad
    K = \text{LayerNorm}(K),
\end{equation}
\begin{equation}
    (\alpha, \beta, \gamma) = \text{Linear}(t_{\text{emb}}), \quad
    V = \text{LayerNorm}(V) \cdot (1 + \alpha) + \beta
\end{equation}
\begin{equation}
    O = \text{Concat}_{i=1}^H \left( \text{Attention}(Q W_i^Q, K W_i^K, V W_i^V) \right)
\end{equation}
where $H$ is the number of attention heads, \( W_i^Q, W_i^K, W_i^V \in \mathbb{R}^{d \times d_{\text{head}}} \)

The final output is calculated with residual connection as below:
\begin{equation}
    O = O + \gamma \cdot act(OW^O)
\end{equation}
where \( W^O \in \mathbb{R}^{H \cdot d_{\text{head}} \times d} \), and $act$ denotes a non-linear activation function.

We apply the matching module at each layer of the 12 encoding blocks and the mid-block in the UNet backbone of the diffusion model, and each attention module uses 8 heads.

\subsection{Meta-training Settings}
The below setting is for the UNet backbone. The setting for the DiT backbone can be found in Section~\ref{supp:dit}.

\begin{itemize}[leftmargin=*]
    \item \textbf{Checkpoint}: We use the Stable Diffusion v1.5 checkpoint available from the HuggingFace~\cite{stable_diffusion_v1_5}.
    \item \textbf{Hyperparameters}: We train using the AdamW optimizer \cite{adamw} with a learning rate of $1 \times 10^{-5}$ and a weight decay of 0.01. For each training batch, we randomly select two tasks per batch, each accompanied by a support set of three example pairs sampled for its query condition.
    \item \textbf{Spatial condition representation}: Following ControlNet \cite{zhang2023adding}, we represent all conditioning inputs as RGB images with a resolution of 512×512.
\end{itemize}

\subsection{Few-shot Fine-tuning Settings}
When fine-tuning our meta-trained model on 30 support examples of the novel condition type, at most 600 fine-tuning steps with a batch size of 10 (approximately 1 hour on an RTX 3090 GPU) suffice for all six tasks. Early-stopping with a held-out validation example to monitor the denoising can be applied to avoid overfitting. We found that low-level conditions (e.g., edges) tend to converge faster than high-level conditions (e.g., pose). We also adopt AdamW optimizer \cite{adamw} with a learning rate of $1 \times 10^{-5}$ and a weight decay of 0.01.

\subsection{Computation Resources}
\begin{table}[h]
\begin{tabular}{lcccc}
\hline
{ } & { GPU(s)} & { Batch size/GPU} & { Mem/GPU} & { Time} \\ \hline
Training                & 8 RTX 3090                    & 6                                     & 16GB                           & 12 hours                    \\
Fine-tuning             & 1 RTX 3090                    & 10                                    & 21GB                           & \textless{}= 1 hour         \\
Inference               & 1 RTX 3090                    & 8                                     & 11.5GB                         & 3.06 s / image            \\ \hline
\end{tabular}
\end{table}

\subsection{Dataset Construction}
\begin{itemize}[leftmargin=*]
    \item \textbf{Training data}: We randomly sample a subset of data from the LAION dataset \cite{schuhmann2021laion}. To identify images containing humans, we use the YOLO11x model \cite{Jocher_Ultralytics_YOLO_2023}. Based on this filtering, we construct a dataset consisting of 150K images with humans and 150K images without humans.
    \item \textbf{Evaluation data}: For the Canny, HED, Depth, and Normal tasks, we evaluate our model on 5,000 images from the COCO2017 validation set \cite{coco2017}. For the Pose task, evaluation is limited to images where humans are successfully detected by the Openpose model \cite{openpose}. Similarly, for the Densepose task, we only evaluate our model with images where the Densepose model \cite{guler2018densepose} detects a human.
\end{itemize}


\section{Comparison with Training-Free Methods}
\vspace{-0.05in}
\label{supp:baselines}
We propose a few-shot framework for adapting to new spatial conditions in T2I diffusion models. A natural question arises: \textit{why use a few-shot approach when training-free methods exist \cite{ugd, yu2023freedom, mo2024freecontrol, ctrl_x}}. To illustrate the advantages of our method (UNet backbone), as mentioned earlier in Section~\ref{sec:exp_setup}, we compare it with Ctrl-X~\cite{ctrl_x} and FreeControl \cite{mo2024freecontrol} state-of-the-art training-free approaches capable of handling diverse spatial conditions without relying on additional pretrained networks. 

To perform spatial control using a condition image, Ctrl-X~\cite{ctrl_x} replaces the features and attention maps when processing the noisy latent with the ones encoded from the noisy condition image. We re-implement Ctrl-X using the Stable Diffusion v1.5 backbone for a comparable setting. FreeControl~\cite{mo2024freecontrol} controls image structure by constructing a PCA basis from object features, projecting both the condition and noisy image feature maps onto this basis, and minimizing an energy function that encourages alignment between the two projections during generation.




\vspace{-0.05in}
\paragraph{Evaluation protocol}
We follow FreeControl’s original evaluation protocol and assess controllability of UFC and two training-free baselines on 30 images from the ImageNet-R-TI2I dataset \cite{tumanyan2023plug}, which includes 10 object categories with three captions each. FreeControl has limited flexibility in generating diverse object categories, as it requires constructing a separate PCA basis for each. This makes large-scale evaluation on 5,000 images from the COCO 2017 validation set \cite{coco2017} impractical, since each prompt must be individually inspected to identify objects and build corresponding PCA bases.
 
As the evaluation dataset of FreeControl excludes human subjects, we compare UFC (30-shot) against FreeControl on four tasks: Canny, HED, Depth, and Normal. Due to the limited number of images, we do not report FID \cite{fid} or Inception Score \cite{inceptionscore}, as they would not provide reliable estimates of image quality.
\vspace{-0.05in}
\paragraph{Result comparison}
Table~\ref{table:freecontrol_compare} shows that UFC (\textit{30-shot}) significantly outperforms both training-free baselines in controllability across all four tasks. Moreover, on a single NVIDIA RTX 3090 GPU, FreeControl takes approximately 100 times longer per image to generate compared to UFC when generating 30 images for evaluation. This overhead stems from the need to construct a PCA basis for each new object category and perform 200 denoising steps for latent optimization. Ctrl-X requires self-recurrence iteration to avoid out-of-distribution sampling, which increases the generation time for several seconds.  In contrast, UFC only introduces a negligible time increase.

\begin{table}[h]
\caption{Controllability scores and average generation time of training-free baselines and UFC (Ours, \textit{30-shot}) on generating 30 images from ImageNet-R-TI2I~\cite{tumanyan2023plug}.}
\label{table:freecontrol_compare}
\centering
\footnotesize
\begin{tabular}{@{}l|cccc|cc@{}}
\toprule
\multirow{2}{*}{Method} & Canny & HED & Depth & Normal & Time\\
 & SSIM ($\uparrow$) & SSIM ($\uparrow$) & MSE ($\downarrow$) & MAE ($\downarrow$) & (Second) \\ \midrule
FreeControl~\cite{mo2024freecontrol} & 0.3139 & 0.3821 & 97.36 & 19.68 & 251.3 \\

Ctrl-X~\cite{ctrl_x} & 0.3896 & 0.3693 & 96.12 & 20.84 & 8.01 \\

\textbf{UFC (Ours)} & \textbf{0.4074} & \textbf{0.5718} & \textbf{93.09} & \textbf{17.75} & 3.06 &\\ \bottomrule
\end{tabular}
\end{table}

\begin{figure}[t]
  \centering
  \vspace{-0.15in}
  \hspace{-0.65in}
  \includegraphics[width=\linewidth]{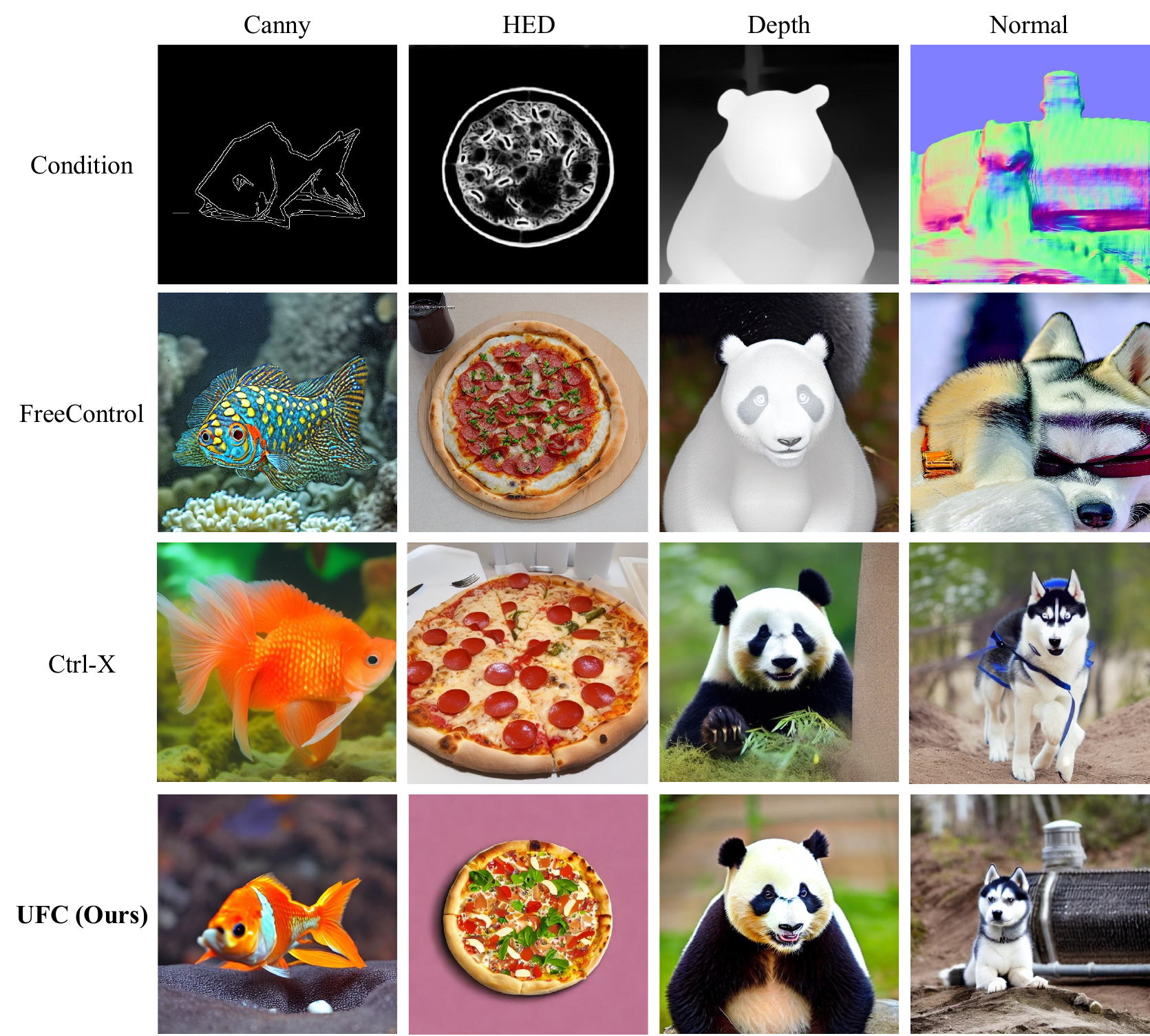}
  \vspace{-0.05in}
  \caption{Qualitative comparison between UFC (\textit{30-shot}) and FreeControl~\cite{mo2024freecontrol}, Ctrl-X~\cite{ctrl_x} on four control tasks.
    }
    \label{fig:freecontrol_compare}
\end{figure}

Figure~\ref{fig:freecontrol_compare} presents qualitative comparisons. UFC accurately follows the spatial condition, while FreeControl and Ctrl-X fail on fine details control.

\clearpage

\section{More Results on Analysis}
\label{supp:shots}
\begin{figure}[h]
  \centering
  \includegraphics[width=0.9\linewidth]{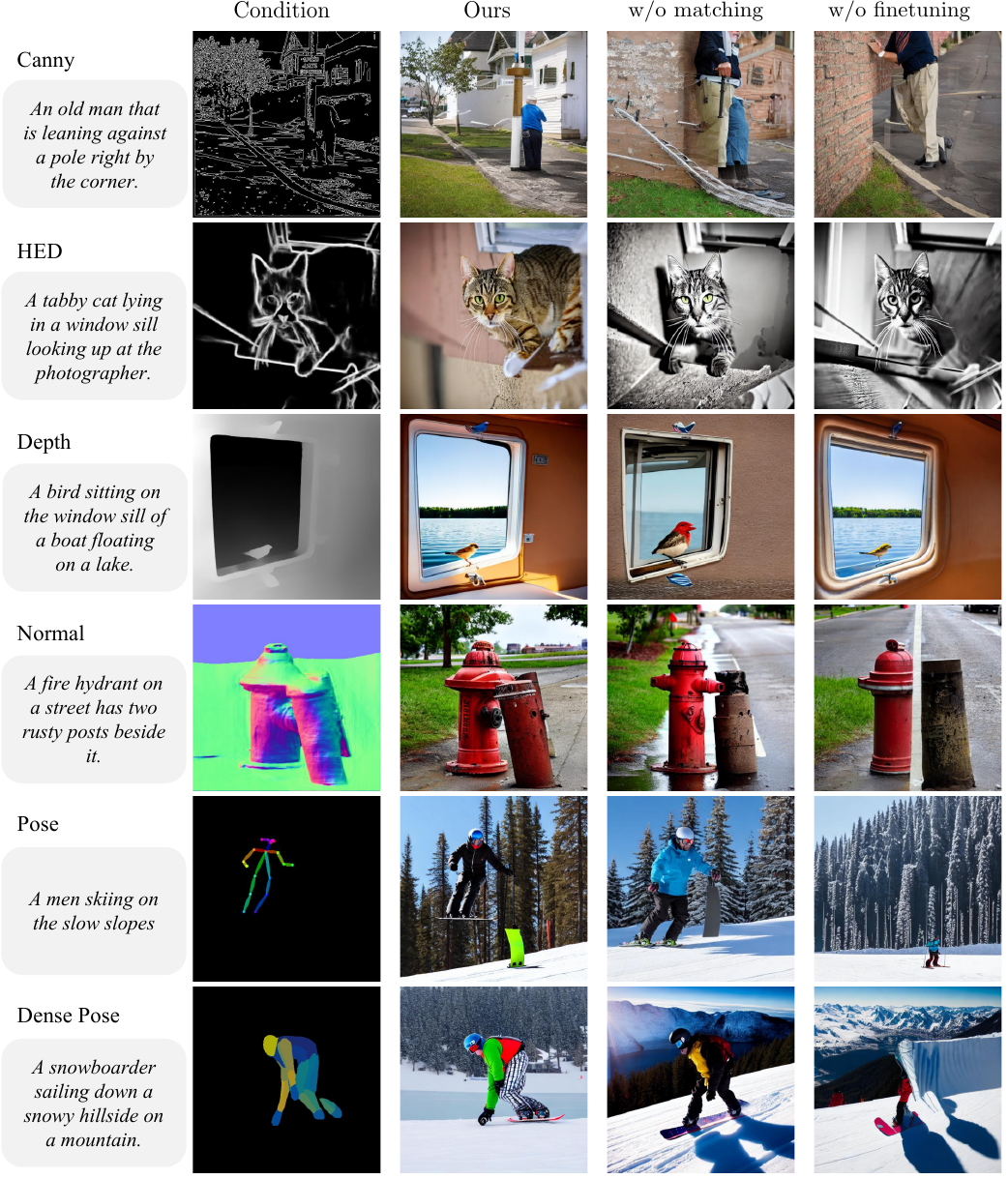}
  \caption{Qualitative comparison results of our method and its two variants: (1) UFC w\textbackslash o matching and (2) UFC w\textbackslash o fine-tuning.
    }
    \label{supp:fig:comparison_variants}
\end{figure}

\paragraph{Qualitative results with two variants}
As mentioned in the ablation study in Section~\ref{sec:analysis},
Figure~\ref{supp:fig:comparison_variants} presents a qualitative comparison of our method with its two variants: (1) UFC w\textbackslash o Matching and (2) UFC w\textbackslash o fine-tuning.
Both variants exhibit degraded image generation in terms of controllability, as they often only partially adhere to the spatial conditions.
For instance, in the 2nd row with HED condition, the generated images from UFC w\textbackslash o Matching and UFC w\textbackslash o Fine-tuning capture the cat's head structure but fail to adhere to the condition for the cat's legs and body. 
In contrast, our full method, which include both matching and fine-tuning, consistently follows all given conditions.

\begin{figure}[h]
  \centering
  \includegraphics[width=0.9\linewidth]{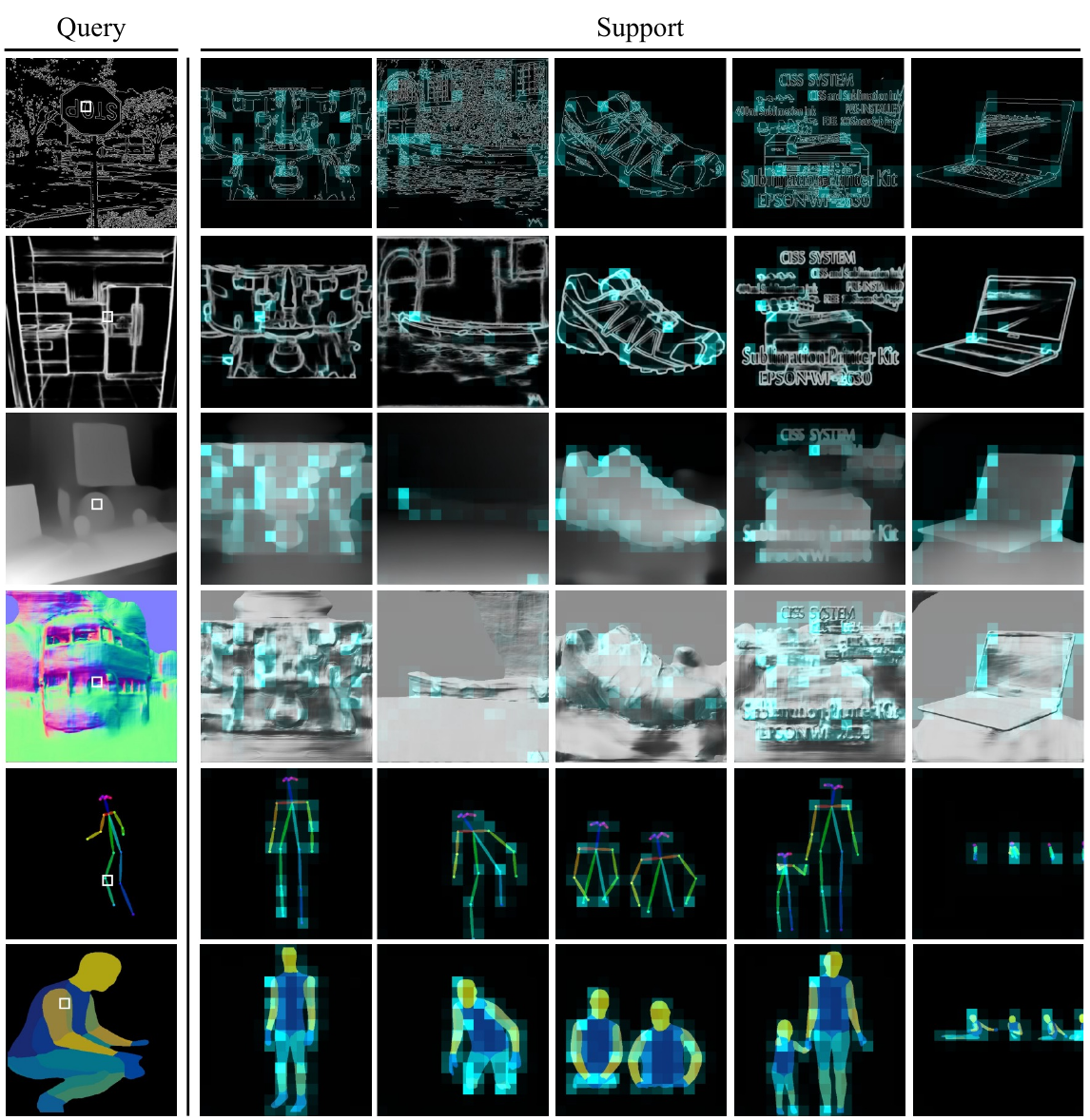}
  \caption{Attention Maps of UFC (\textit{30-shot}) from the 7th layer and a selected head for each task. Support normal maps are converted to grayscale to enhance the visibility of attention regions. Query patches attend primarily to the most relevant support patches.
    }
    \label{supp:fig:attn_map}
    \vspace{-0.1in}
\end{figure}

\paragraph{Attention map visualization} 
We present an example of attention maps in Figure~\ref{supp:fig:attn_map}. For each task, the selected query patch (highlighted by a white box in the Query column) can attend to relevant support patches, rather than unrelated regions such as background areas. For instance, in the Canny and HED tasks (first two rows), the query patches focus on support regions that preserve similar edge structures. On the other hand, for the Pose and Densepose tasks (last two rows), the query patches attend to regions related to human body parts.
\vspace{-0.1in}

\begin{figure}[h]
  \centering
  \includegraphics[width=\linewidth]{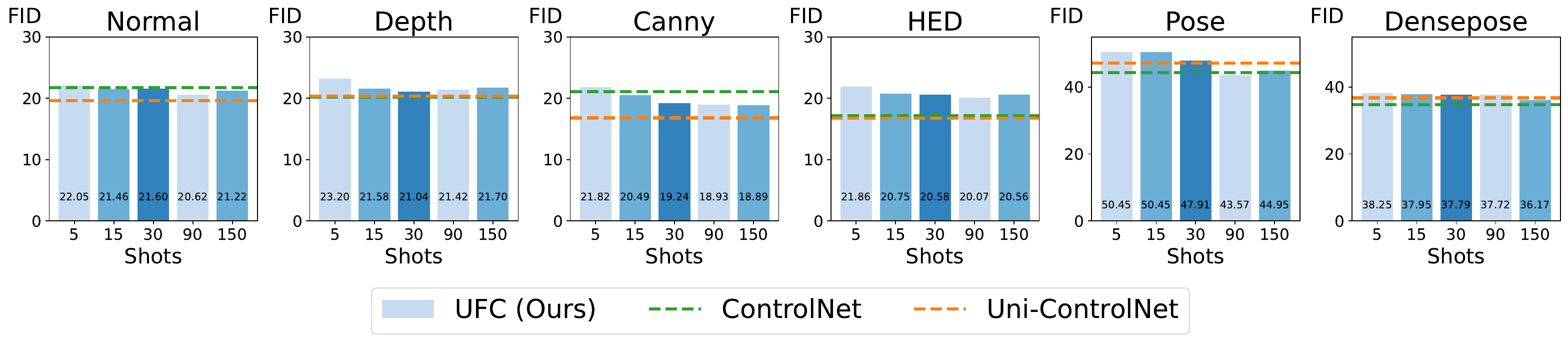}
  \vspace{-0.2in}
  \caption{FID score over varying support set sizes (shots).}
  \label{supp:fig:fid}
\end{figure}

\paragraph{FID Results with different number of shots}
In addition to the controllability measurement in Figure~\ref{fig:numshots}, Section~\ref{sec:analysis}, we provide FID results obtained by fine-tuning our method (UFC) using different numbers of support data (i.e., shots) in 
Figure~\ref{supp:fig:fid}.
The results show that our method maintains FID scores across various shots.
Specifically, for Normal, Depth, Canny, and HED tasks, our method preserves image quality with FID changes remaining within 1.
For Pose, the FID difference is within 4.5, and for DensePose, it is within 1.7.
Both changes are relatively small given the FID scores of each task, and using more support data often slightly improves FID. The results confirm that our method improves controllability as the number of support data increases, while maintaining image quality.

\paragraph{Effect of support samples}
As discussed in Section~\ref{sec:analysis}, UFC is evaluated in 30-shot settings with different investigate the effect of support sets' diversity. In this section, we would like to discuss the implications of \textit{diversity} for each task. Because our matching strategy hinges on condition-specific similarity, the notion of \textit{diversity} depends on tasks. For edge inputs such as Canny and HED, the support set must span a wide range of edge-map densities and orientations. For Pose and DensePose, it needs to cover varied human scales, occlusions, and pose deformations.
To measure how support choice influences performance, we repeated fine-tuning three times, each time drawing a new random set of 30 supports and carrying out few-shot generation. The results are summarized in the Table~\ref{table_diversity}.
\begin{table}[h]
\caption{Analysis on the effect of support set's diversity on model performance.}
\label{table_diversity}
\footnotesize
\centering
\begingroup            
\setlength{\tabcolsep}{3pt}   
\begin{tabular}{@{}l*{12}{c}@{}}   
\toprule
\multirow{2}{*}{seed} & \multicolumn{2}{c}{Canny} & \multicolumn{2}{c}{HED}
                         & \multicolumn{2}{c}{Depth} & \multicolumn{2}{c}{Normal}
                         & \multicolumn{2}{c}{Pose} & \multicolumn{2}{c}{Densepose} \\
\cmidrule(l){2-3} \cmidrule(l){4-5} \cmidrule(l){6-7}
\cmidrule(l){8-9} \cmidrule(l){10-11} \cmidrule(l){12-13}
 & SSIM$\uparrow$ & FID & SSIM$\uparrow$ & FID
 & MSE$\downarrow$ & FID & MAE$\downarrow$ & FID
 & $\text{AP}^{50}$$\uparrow$ & FID & mIoU$\uparrow$ & FID \\ \midrule
1               & 0.3239 & 19.24 & 0.5121 & 20.58
                    & 94.38  & 21.04 & 15.09  & 21.60
                    & 0.229  & 47.91 & 0.4340 & 37.79 \\
2
                  & 0.3295 & 21.06 & 0.4957 & 20.70
                    & 93.47  & 20.51 & 15.82  & 20.85
                    & 0.176 & 48.22 & 0.3827 & 39.52  \\
3
                    & 0.3273 & 20.90 & 0.5128 & 23.84
                    & 94.84  & 22.21 & 15.93  & 22.56
                    & 0.191 & 48.66 & 0.3831 & 37.46 \\
4
                    & 0.3290 & 20.41 & 0.4942 & 20.78
                    & 94.50  & 21.80 & 15.56  & 21.89
                    & 0.202 & 47.83 & 0.3914 & 38.30 \\ \midrule
                    
Mean               & 0.3274 & 20.40 & 0.5037 & 21.73
                    & 94.30  & 21.39 & 15.60  & 21.73
                    &0.1995 & 48.16 & 0.3978 & 38.27 \\
Std               & 0.0025 & 0.82 & 0.0101 & 1.49
                    & 0.5851  & 0.76 & 0.3737  & 0.71
                    & 0.0224 & 0.38 & 0.0245 & 0.90 \\ \bottomrule
\end{tabular}
\endgroup          
\end{table}

\clearpage

\section{Experiments with DiT Backbone}
\label{supp:dit}

\paragraph{Model}
As discussed in Section~\ref{sec:architecture}, our framework can work with both UNet and DiT backbones. We extend UFC to the DiT architecture~\cite{dit}, using Stable Diffusion v3~\cite{sdv3} as an example, initialized from the publicly available v3.5-medium checkpoint~\cite{stablediffusion3.5medium} on Hugging Face. This model consists of 24 Transformer layers in the diffusion backbone. We initialize the image encoder with pretrained weights of the diffusion backbone and keep it frozen during training. For the condition encoder, we initialize it as a trainable copy of the first 12 layers of the diffusion backbone. We train our model on 8 NVIDIA RTX A6000 GPUs.

\paragraph{Control feature injection}
Unlike the UNet backbone, the DiT architecture does not have skip connections. Therefore, we inject the output of the matching module directly into the hidden representations at each layer of the DiT backbone. Specifically, image features are extracted from 12 even-numbered layers of the image encoder, while query and support condition embeddings are obtained from the condition encoder. The matching mechanism is applied across these 12 layers to generate multi-layer control features. The 24-layer DiT diffusion backbone is divided into 12 sequential chunks, each containing 2 layers. The control features are then added to every hidden representation of these chunks in the corresponding order. 

\paragraph{Hyper-parameters}
We follow the same optimizer settings, training dataset, and evaluation protocol described in Section ~\ref{sec:exp_setup}. For inference, we use the flow-matching Euler scheduler introduced in Stable Diffusion 3\cite{sdv3}, with a classifier-free guidance (CFG) scale of 5.0, 28 generation steps, and a fixed seed of 42.

\begin{figure}[h]
  \centering
  \includegraphics[width=\linewidth]{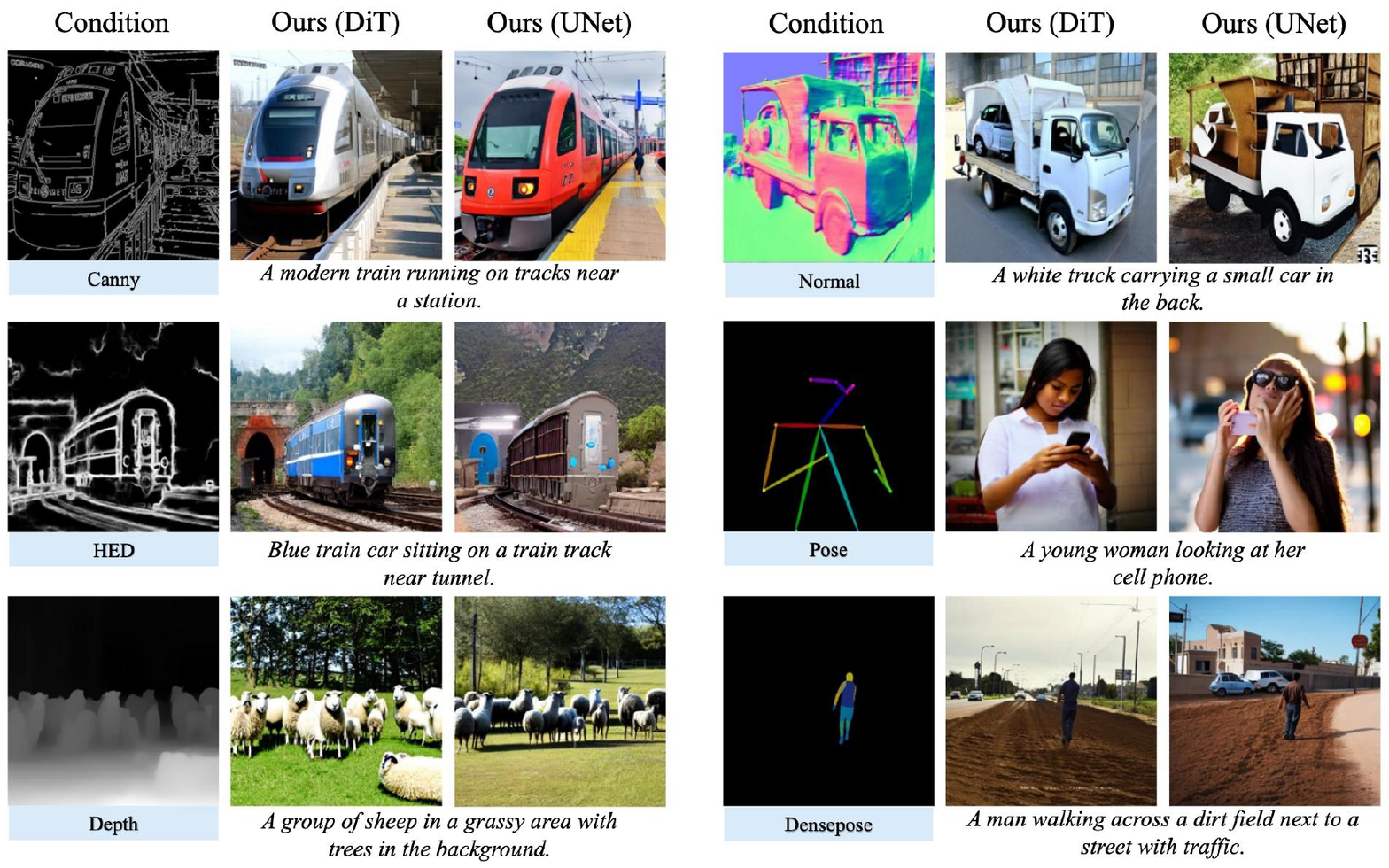}
  \vspace{-0.4in}
  \caption{
    Qualitative comparison of UFC using DiT and UNet backbones in \textit{30-shot} setting. Our method with the DiT backbone yields more fine-grained spatial control than the UNet counterpart.
    }
    \label{fig:DiT_UNet_compare}
\end{figure}

\begin{figure}[t]
  \centering
  \hspace{-0.42in}
  \includegraphics[width=\linewidth]{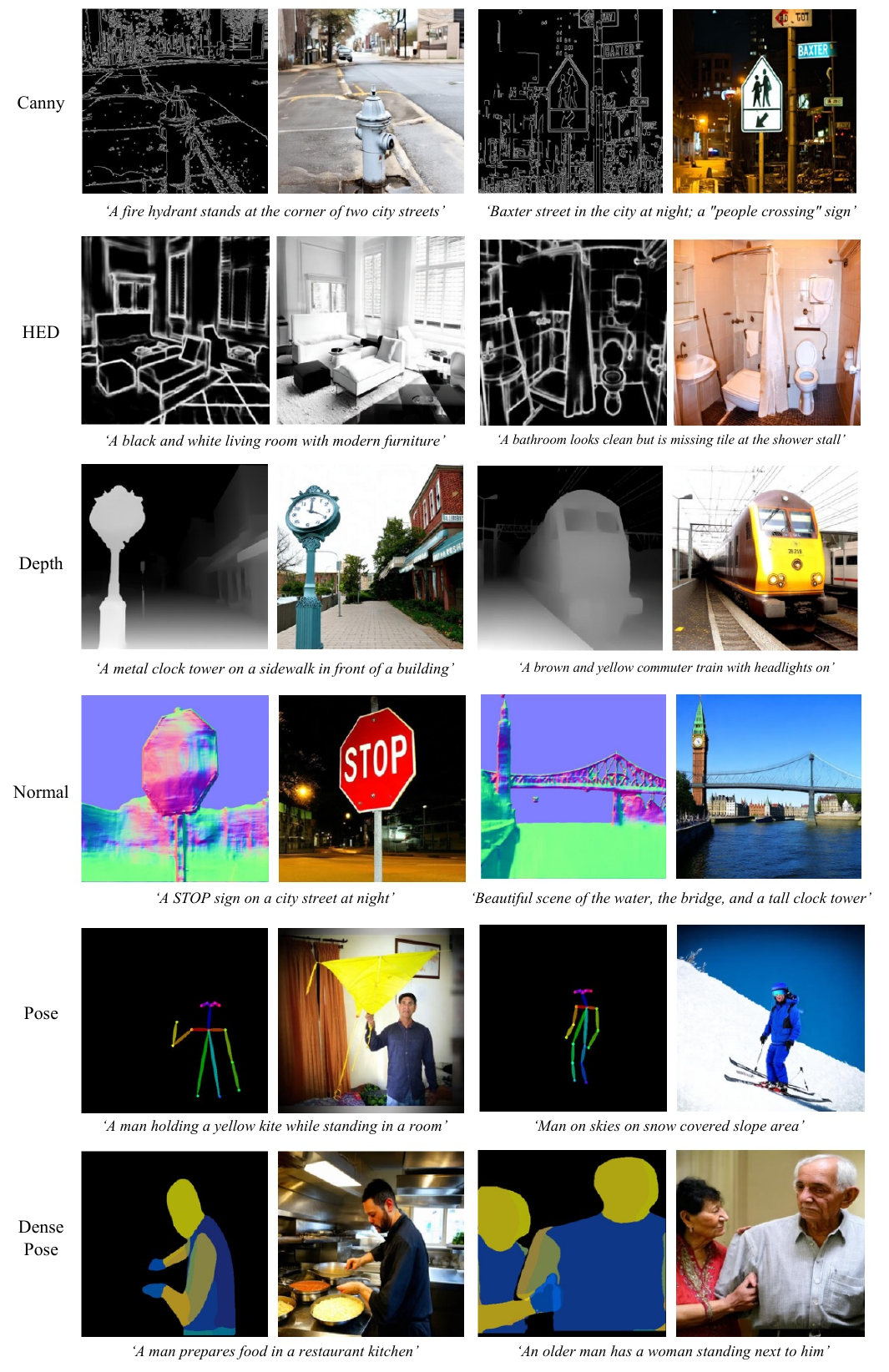}
  \caption{Generated images from UFC with DiT backbone in \textit{30-shot} setting.
    }
    \label{supp:fig:dit}
\end{figure}

\paragraph{Qualitative Results}
Qualitative examples using the DiT backbone are shown in Figures~\ref{fig:DiT_UNet_compare},~\ref{supp:fig:dit}. Our DiT-based model, adapted with only \textbf{30 shots}, is able to control the structure of generated images with various novel condition tasks.

\clearpage

\section{More Results}
\label{supp:more_results}
\paragraph{Results with more spatial conditions}

\begin{figure}[h]
    \centering
  \hspace{-0.42in}
    \includegraphics[width=\linewidth]{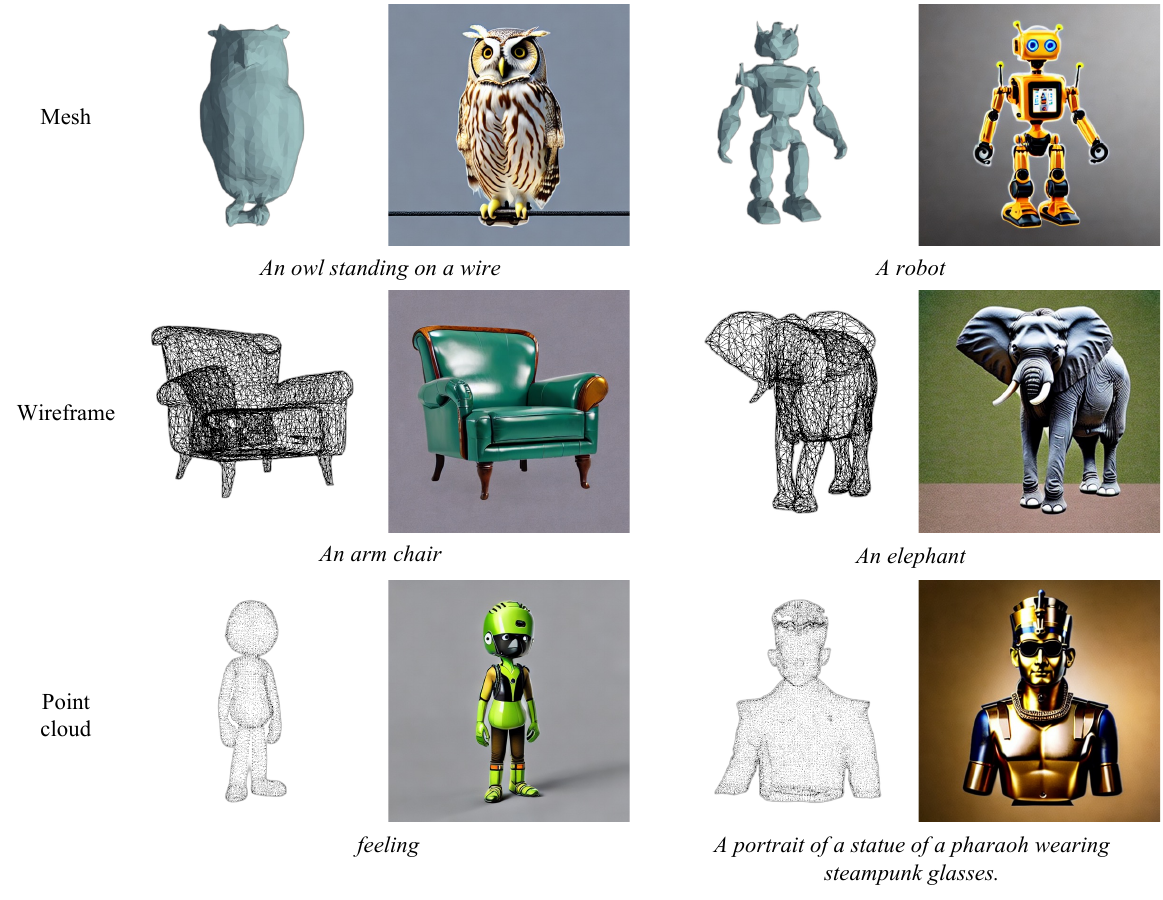}
    \caption{Generated images from UFC with spatial conditions of 3D meshes, wireframes, and point clouds in \textit{30-shot} setting.}
    \label{fig:supp:new_condition}
\end{figure}

As previously discussed in Section~\ref{sec:analysis}, we evaluate UFC on more novel spatial conditions, including 3D meshes, wireframes, and point clouds. We use iso3d~\cite{ebert20253d}, a 3D isolated object dataset(with no background), to generate spatial conditions for image pairs.  
We only report qualitative results, as the absence of a pre-trained condition-prediction network prevents us from measuring controllability using the evaluation approach in Section~\ref{sec:exp_setup}. 
Moreover, the insufficient collected validation sets hinder reliable assessment of image quality with FID~\cite{fid} or Inception Score~\cite{inceptionscore}.
Figure~\ref{fig:supp:new_condition} shows the qualitative results of our method, demonstrating the effectiveness of our method in a few-shot (30 supports) setting.

\begin{figure}[t]
  \centering
  \hspace{-0.42in}
  \includegraphics[width=\linewidth]{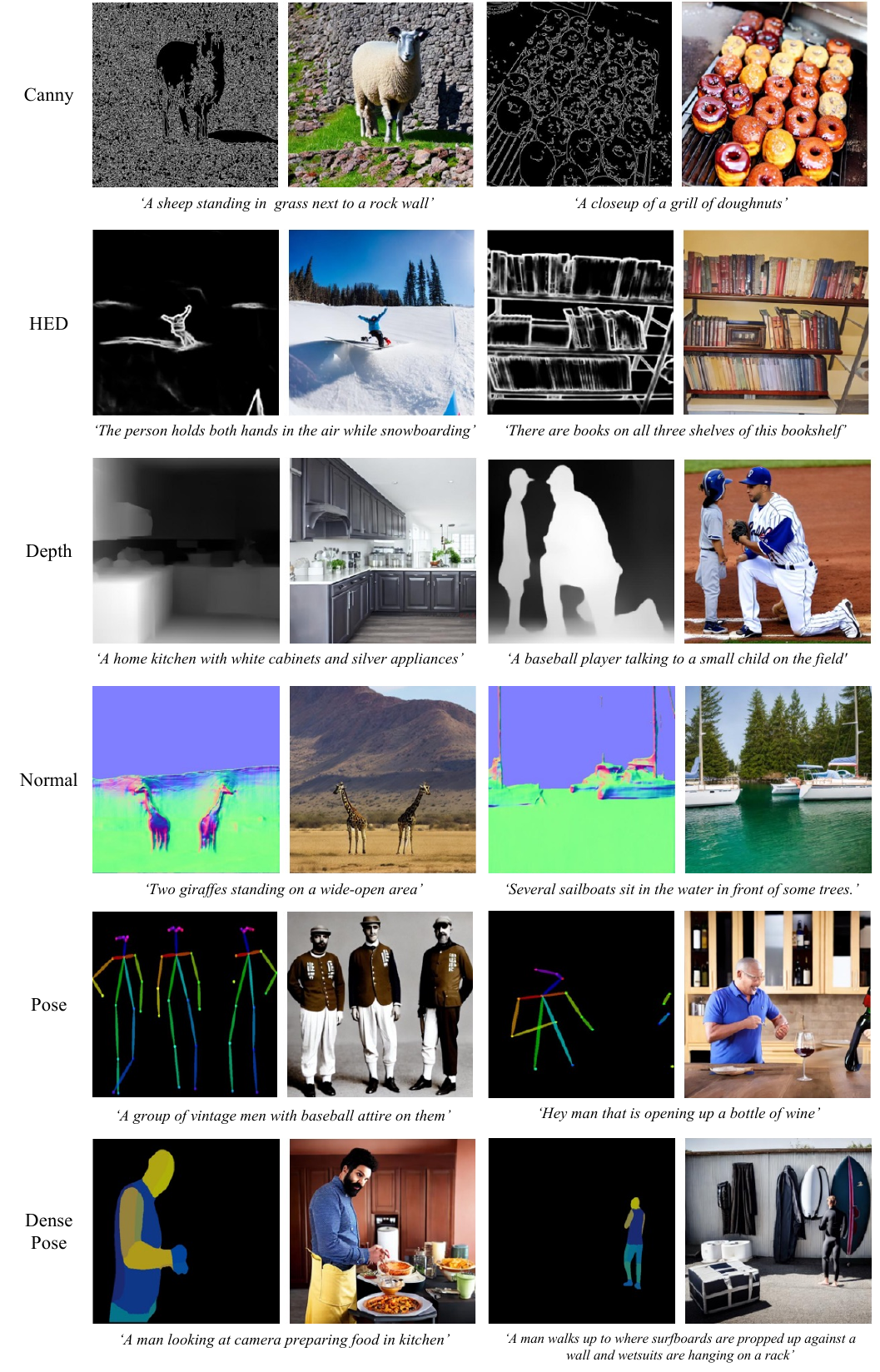}
  \caption{More qualitative results of UFC (UNet) in \textit{30-shot} setting.}
    \label{supp:fig:canny_hed_depth}
\end{figure}

\begin{figure}[t]
  \centering
  \hspace{-0.42in}
  \includegraphics[width=\linewidth]{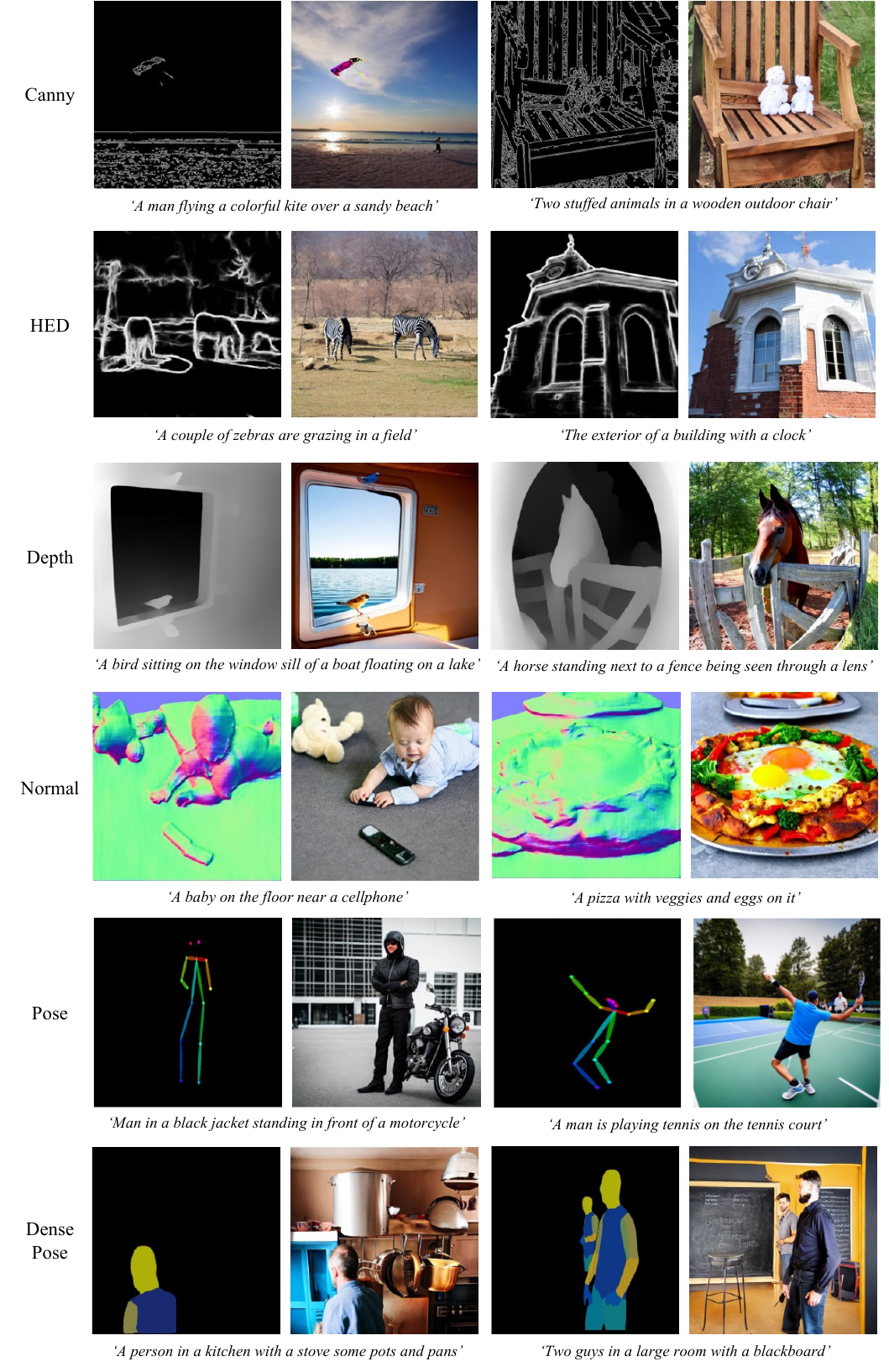}
  \caption{More qualitative results of UFC (UNet) in \textit{30-shot} setting.}
    \label{supp:fig:normal_pose_densepose}
\end{figure}

\paragraph{More qualitative results}
We present more qualitative results of UFC (UNet) in Figure~\ref{supp:fig:canny_hed_depth} and Figure~\ref{supp:fig:normal_pose_densepose}. All the results are generated with a support size of 30.

\clearpage

\clearpage

\section{Support Image-Condition Pairs}\label{supp:supp_pairs}
As described in implementation detail in Section~\ref{sec:exp_setup}, after few-shot fine-tuning, 5 image–condition pairs are used for inference.

\begin{figure}[h]
  \centering
  \begin{minipage}{\linewidth}
    \centering
  \hspace{-0.42in}
    \includegraphics[width=0.9\linewidth]{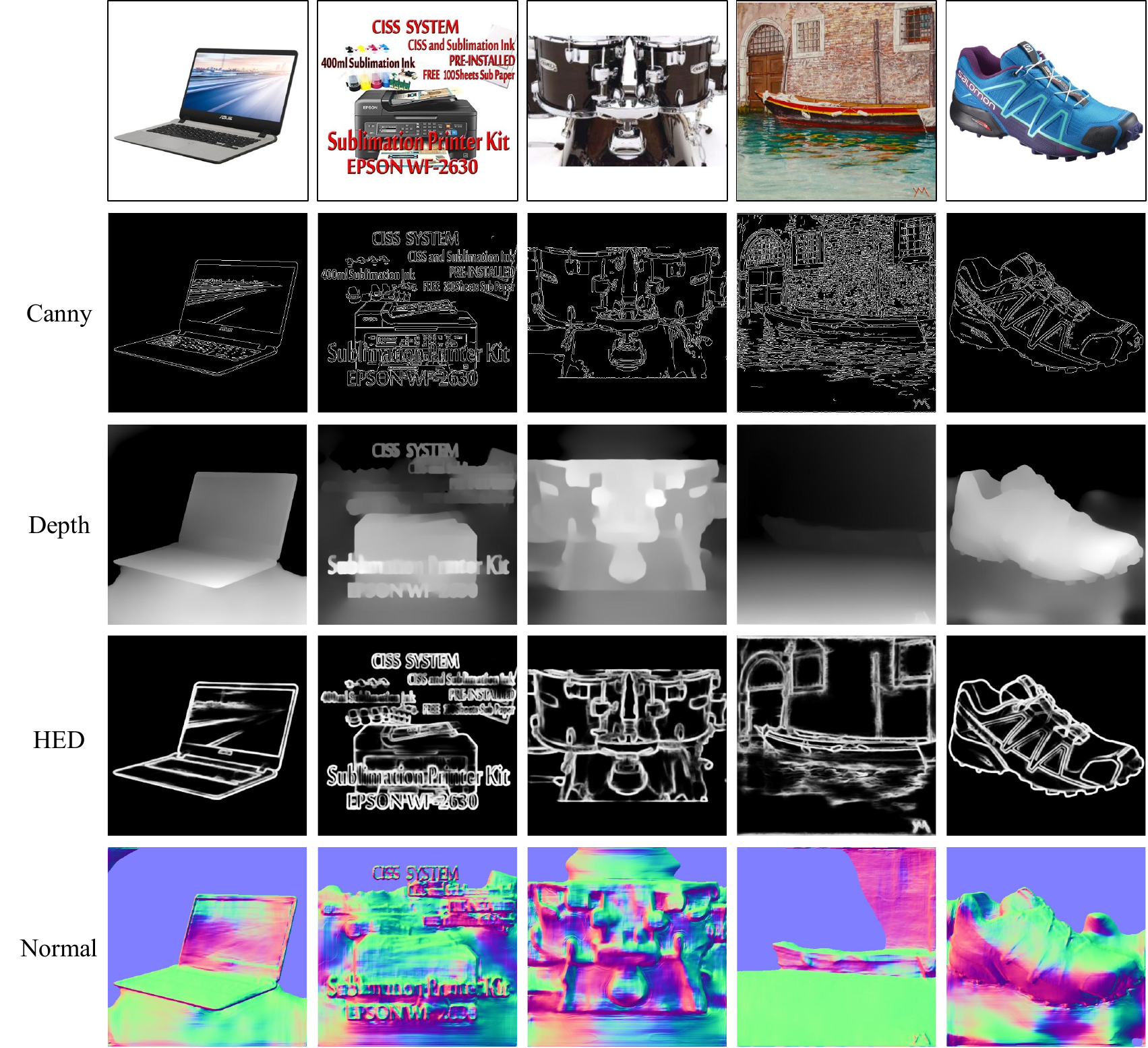}
    \vspace{-0.05in}
    \caption{Five support image-label pairs used for evaluating Canny/Depth/HED/Normal tasks.}
    \label{supp:fig:support_non_human}
    \vspace{0.1in}
  \end{minipage}
  \vspace{1.0em} 
  \begin{minipage}{0.98\linewidth}
    \centering
  \hspace{-0.35in}
    \includegraphics[width=0.9\linewidth]{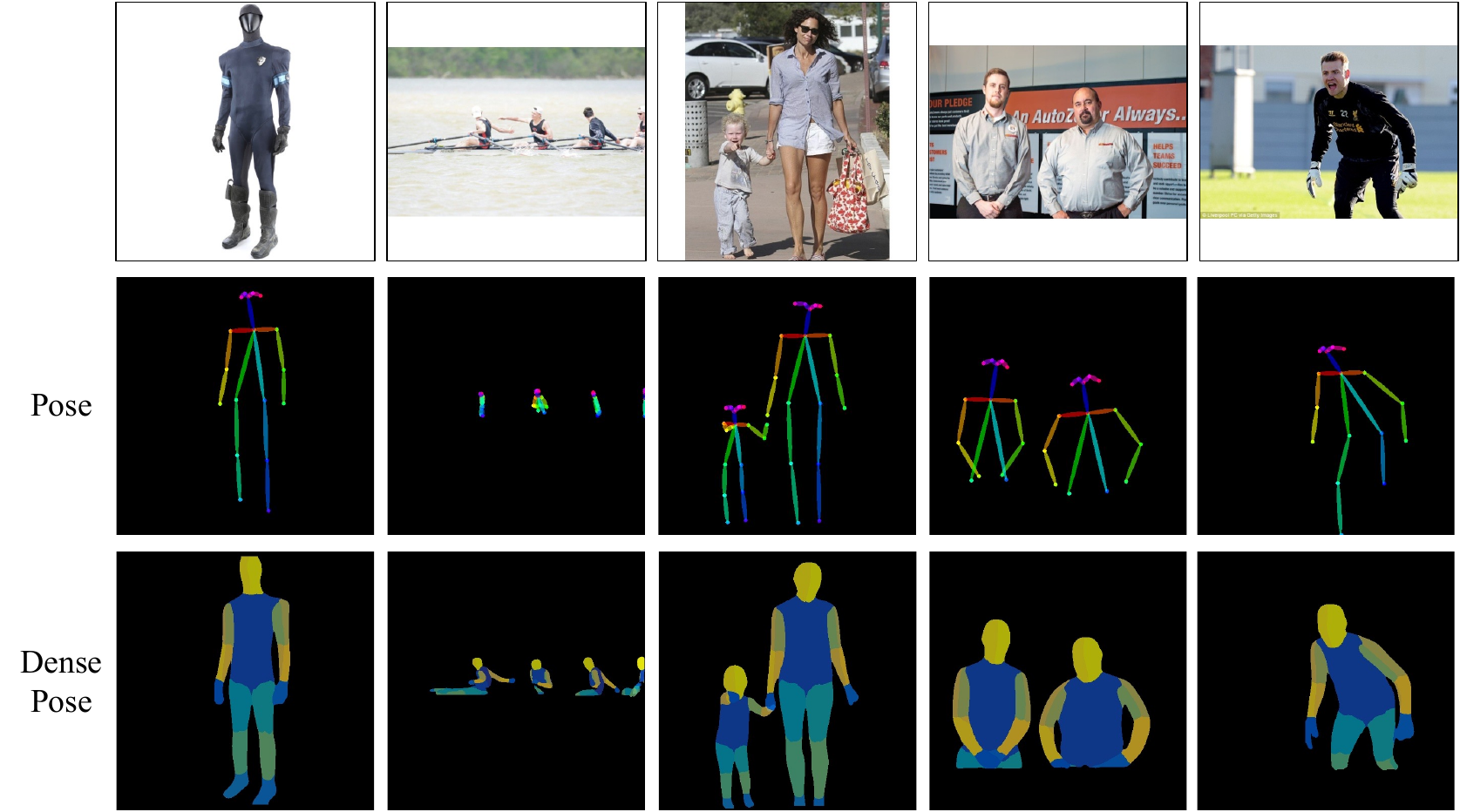}
    \vspace{-0.05in}
    \caption{Five support image-label pairs used for evaluating Pose/DensePose tasks.}
    \label{supp:fig:support_human}
  \end{minipage}
\end{figure}

\end{document}